%% file: main.tex
\documentclass[10pt,twocolumn,letterpaper]{article}

\usepackage[pagenumbers]{cvpr} 

\usepackage{graphicx}
\usepackage{amsmath}
\usepackage{amssymb}
\usepackage{booktabs}

\usepackage{braket}
\usepackage{nicefrac}

\DeclareMathOperator*{\argmin}{arg\,min}
\usepackage{array}
\usepackage{multirow}

\usepackage[pagebackref,breaklinks,colorlinks]{hyperref}

\newcommand{\vct}[1]{\mathbf{#1}}

\usepackage[capitalize]{cleveref}
\crefname{section}{Sec.}{Secs.}
\Crefname{section}{Section}{Sections}
\Crefname{table}{Table}{Tables}
\crefname{table}{Tab.}{Tabs.}

\def\cvprPaperID{****} 
\def\confName{CVPR}
\def\confYear{2022}

\begin{document}

\title{Adiabatic Quantum Computing for Multi Object Tracking}

\author{Jan-Nico~Zaech$^{1}$\quad Alexander Liniger$^{1}$\quad Martin Danelljan$^{1}$\quad Dengxin~Dai$^{1,2}$\quad Luc~Van~Gool$^{1,3}$\\
$^{1}$Computer Vision Laboratory, ETH Zurich, Switzerland,\\
$^{2}$MPI for Informatics, Saarbrucken, Germany,
$^{3}$KU Leuven, Belgium\\
{\tt\small\{zaechj,alex.liniger,martin.denelljan,dai,vangool\}@vision.ee.ethz.ch}}
\maketitle

\begin{abstract}

Multi-Object Tracking (MOT) is most often approached in the tracking-by-detection paradigm, where object detections are associated through time. The association step naturally leads to discrete optimization problems.
As these optimization problems are often NP-hard, they can only be solved exactly for small instances on current hardware.
Adiabatic quantum computing (AQC) offers a solution for this, as it has the potential to provide a considerable speedup on a range of NP-hard optimization problems in the near future. However, current MOT formulations are unsuitable for quantum computing due to their scaling properties.
In this work, we therefore propose the first MOT formulation designed to be solved with AQC.
We employ an Ising model that represents the quantum mechanical system implemented on the AQC. 
We show that our approach is competitive compared with state-of-the-art optimization-based approaches, even when using of-the-shelf integer programming solvers. Finally, we demonstrate that our MOT problem is already solvable on the current generation of real quantum computers for small examples, and analyze the properties of the measured solutions.
\end{abstract}

\begin{figure}[t]
\centering
\includegraphics[width=\linewidth, trim= 300 60 300 60, clip]{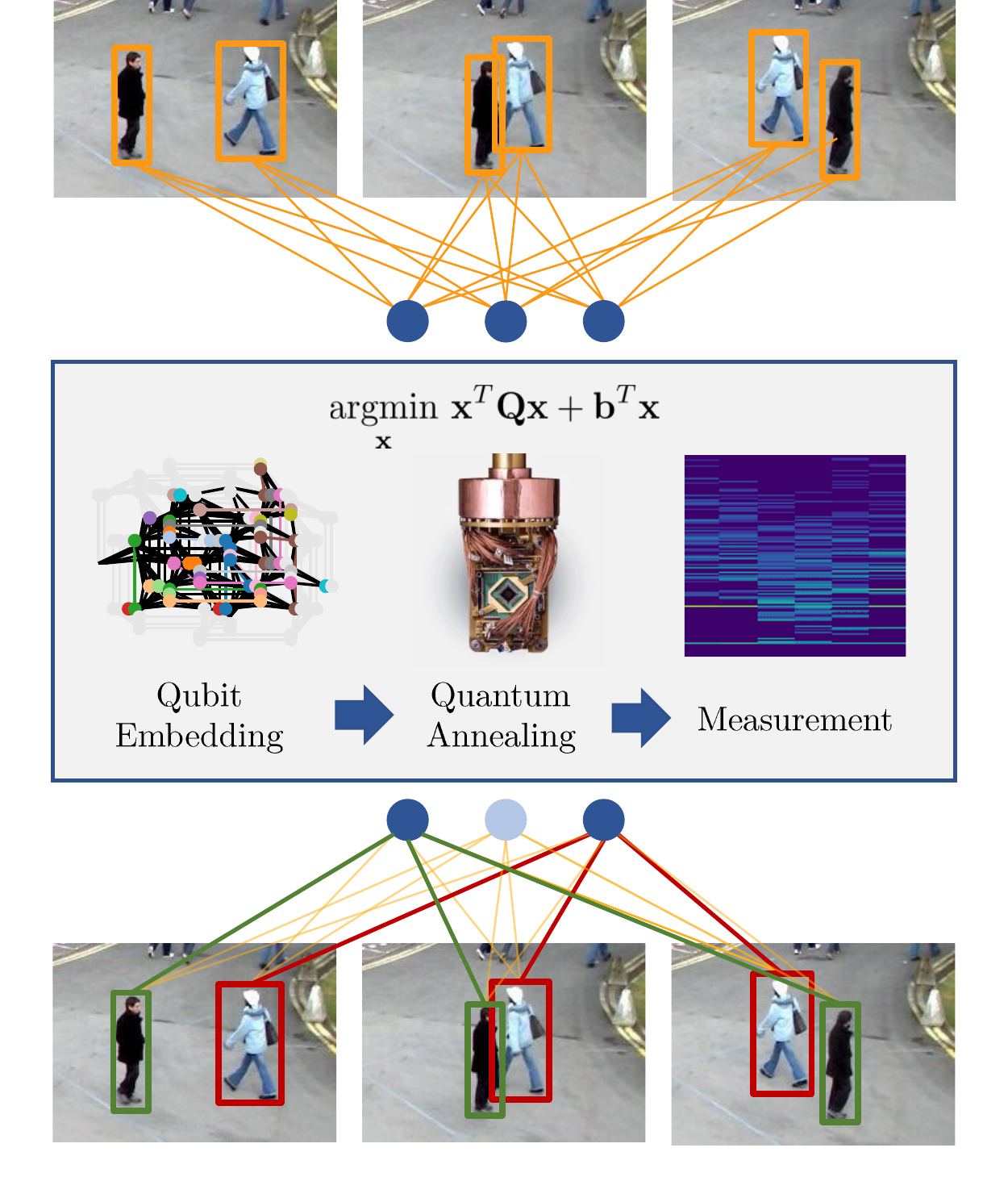}
\label{fig:intro}
\vspace{-10pt}
\caption{The proposed approach to MOT states the assignment problem between detections and a set of tracks as a quadratic unconstrained binary optimization task. We then represent the optimization problem as a quantum mechanical system that can be implemented on an AQC. Via quantum annealing, a minimum energy state is found that represent the best assignment.}
\end{figure}

\section{Introduction}
\label{sec:intro}

Multi-Object Tracking (MOT) is a task in computer vision that requires solving NP-hard assignment problems~\cite{hornakova_lifted_2020, hornakova_making_2021, tang_multiple_2017}. To make this feasible, the community proposed a range of different approaches: work on the problem formulation using domain knowledge helps to make it an easier to solve problem~\cite{hornakova_lifted_2020, tang_multiple_2017}, approximate solvers extend the feasible problem size~\cite{hornakova_making_2021}, and the combination of deep learning with simple heuristics can be seen as a data-driven approach to the problem~\cite{braso_learning_2020, dai_learning_2021}. Nevertheless, integer assignment problems remain hard optimization tasks for any available solver. With the recent progress in quantum computing, a new way of solving such optimization problems becomes feasible in the near future~\cite{king_scaling_2021, apolloni_quantum_1989, van_dam_how_2001}.

Instead of iteratively exploring possible solutions, e.g. via branch and bound, the problem is mapped to a quantum mechanical system, whose energy is equivalent to the cost of the optimization problem. Therefore, if it is possible to measure the lowest energy state of the system, a solution to the corresponding optimization problem is found. This is done with an adiabatic quantum computer (AQC), which implements a quantum mechanical system made from qubits and can be described by the Ising model~\cite{ising_beitrag_1925}. Using this approach, a quantum speedup, which further scales with system size and temperature, has already been shown for applications in physics~\cite{king_observation_2018, king_scaling_2021}.

While there is a range of advantages that quantum computing can provide in the future, mapping a problem to an AQC is not trivial and often requires reformulating the problem from scratch, even for well investigated tasks~\cite{benkner_q-match_2021, birdal_quantum_2021}. On the one hand, the problem needs to be matched to the Ising model, on the other hand, real quantum computers have a very limited number of qubits and are still prone to noise, which requires tuning of the model to handle the limitations.

In this work we present the first quantum computing approach to MOT. The number of required qubits in our formulation grows linearly in the number of detections, tracks and timesteps and only requires entanglement between qubits to model long-term relations. Our overall contributions are the following:
\begin{itemize}
    \setlength{\itemsep}{0pt}
    \setlength{\parskip}{0pt}
    \item A quantum computing formulation of MOT that is competitive with state-of-the-art methods.
    \item A method using few problem measurements to find Lagrange multipliers that considerably improve solution probability.
    \item Extensive MOT experiments on synthetic as well as real data using a D-wave AQC.
\end{itemize}

The remaining paper is structured as follows: After presenting related work, the basics of quantum computing are introduced. This is followed by our MOT formulation that is optimized to run on an AQC. We then show the changes required to make the problem solvable also with the classical computing paradigm. Finally, experiments on a D-wave quantum computer are presented together with results on larger problem instances.

\section{Related Work}
\label{sec:rel_work}

\textbf{Quantum computing} applications have recently started to emerge across a range of fields that rely on discrete optimization, as adiabatic quantum computers have become accessible. The applications include examples such as gene engineering~\cite{fox_mrna_2021}, interaction reconstruction in particle physics~\cite{das_track_2020}, traffic flow optimization~\cite{neukart_traffic_2017}, or route selection in robotics~\cite{ohzeki_control_2019}. In computer vision, discrete optimization is a ubiquitous part of many applications. While these applications frequently rely on heuristics today, quantum computing has the potential to provide an efficient way of directly solving them. In the area of 3D vision, quantum computing has been used by Feld \etal~\cite{feld_optimizing_2018} for optimizing geometry compression. Benkner \etal~\cite{benkner_adiabatic_2020} use adiabatic quantum computing to match 3D shapes and images with permutation matrices and investigate different constraint formulations to optimize the probability of finding a correct solution. By using an iterative approach, the same authors are able to scale the approach up to larger problem instances~\cite{benkner_q-match_2021}. Closest to our work is the contribution of Birdal \etal~\cite{birdal_quantum_2021}. They map the permutation synchronization task to an optimization problem solvable on a quantum computer and show results on small problem instances.

\textbf{Multi-object tracking} describes the problem of tracking all objects belonging to a predefined set of object types in 2D~\cite{dendorfer_mot20_2020, leal-taixe_motchallenge_2015, milan_mot16_2016, manen_pathtrack_2017} or 3D~\cite{Argoverse, geiger_vision_2013, nuscenes2019}. Most competitive trackers follow a tracking by detection approach, where a set of detections is given in every frame and the trackers perform association between frames, interpolation of occlusions, and rejection of false-positive detections. While most approaches use deep learning to generate appearance features~\cite{ristani_performance_2016, wei_person_2018, zheng_joint_2019, zheng_scalable_2015}, two major groups of data assignment approaches exist. The first one maps the matching step to a deep learning task~\cite{braso_learning_2020, zaech_learnable_2021, zhou_tracking_2020, dai_learning_2021} and uses simple heuristics to resolve the remaining inconsistencies. This allows for training the complete pipeline end-to-end, without the direct requirement to define a cost for data association. The second group directly performs data association using discrete optimization algorithms~\cite{hornakova_lifted_2020, hornakova_making_2021, li_global_2008, roshan_zamir_gmcp-tracker_2012, tang_multiple_2017, ren_tracking-by-counting_2021}, which is stated as a network flow optimization problem in most cases.
These formulations allow to include long-term relations~\cite{hornakova_lifted_2020}, and prior information about the nature of tracks in an intuitive and transparent way. Nevertheless, these properties come at a high computational cost. As most of the proposed optimization problems are NP-hard~\cite{ganian_complexity_2021}, a considerable effort was invested in finding heuristics and approximate solvers for them~\cite{hornakova_making_2021}.

\section{Preliminaries on Quantum Computing}
\label{sec:adiabatic_comp}
Quantum computers are systems operating in a state that is described by its quantum properties, such as superposition and entanglement. By exploiting these properties, a range of problems that quickly grow in complexity on classical computers and thus, cannot be solved in any reasonable timeframe, could be solved considerably faster~\cite{feynman_simulating_1982} by a quantum computer. Reaching such a point is widely referred to as quantum primacy. Even though implementations of quantum computers are still heavily experimental, some problems have already been shown to profit from them, including the sampling of pseudo-random quantum circuits~\cite{arute_quantum_2019, wu_strong_2021} and Gaussian boson sampling \cite{zhong_phase-programmable_2021}.

\noindent\textbf{Qubits} are two-state quantum-mechanical systems that form the basis of quantum computers. Like a bit, a qubit has two basis states $\ket{0} = [ 1~0 ]^T$ and $\ket{1} = [ 0~1 ]^T$ that in a superposition form the qubit's state. Qubits can be realized with a wide range of approaches, including superconducting circuits, ions trapped in an electromagnetic field, or photons.

\noindent\textbf{Quantum Superposition} refers to the property of a quantum system that it is not required to be in one of the basis states, but rather can be described by a linear combination of possible basis states. A qubit in a pure state $\ket{\psi}$ can be described with its two basis states $\{\ket{0}, \ket{1}\}$ as
\begin{equation}
    \ket{\psi} = c_1\ket{0} + c_2\ket{1}
\end{equation}
where $c_1$ and $c_2$ are complex numbers, called probability amplitudes, with $|c_1|^2 + |c_2|^2 = 1$.

\noindent\textbf{Measurement} of a qubit state results in one of the basis states $\{\ket{0}, \ket{1}\}$. The probability of measuring $\ket{0}$ and $\ket{1}$ evaluates to $|c_1|^2$ and $|c_2|^2$, respectively \cite{holevo_repeated_2001}. As a measurement corresponds to an observation of the qubit it leads to wave function collapse, which means that the qubit state is changed irreversibly \cite{griffiths_schroeter_2018}.

\noindent\textbf{Entanglement} of qubits is at the very heart of quantum computing~\cite{jozsa_role_2003}. A system of entangled qubits is represented by a system state where each qubit cannot be described only with its own state but depends on the state of the remaining system~\cite{einstein_can_1935, schrodinger_discussion_1935, kocher_polarization_1967, hensen_loophole-free_2015, ma_quantum_2012}. Thus, measuring a single qubit can collapse the wave function of other entangled qubits, which alters their state and therefore, also their measurement outcome \cite{griffiths_schroeter_2018}.

\subsection{Adiabatic Quantum Computing}
\label{sec:aqc}
Adiabatic quantum computing~\cite{farhi_quantum_2000, van_dam_how_2001} is an approach that, instead of using gates as unitary operations on subsets of the available qubits~\cite{deutsch_quantum_1989}, uses a problem Hamiltonian $\hat{H}_P$ that describes the operation applied on all qubits simultaneously. The problem Hamiltonian is designed such that its ground-state, which is the lowest energy configuration of the system, represents the result for a computational task~\cite{apolloni_quantum_1989}. In general, a Hamiltonian $\hat{H}(t)$ is an operator that represents the energy of a quantum system and can be used in the Schr\"odinger equation to describe the system's evolution over time as
\begin{equation}
\label{eq:evolution_schrodinger}
i \hbar \frac{\partial}{\partial t} \ket{\psi(t)} = \hat{H}(t)\ket{\psi(t)},
\end{equation}
where $i$ is the imaginary unit, and $\hbar$ the reduced Planck constant.

As the ground-state of the problem Hamiltonian is hard to find, the complete system is initialized with an initial Hamiltonian $\hat{H}_B$ that has an easy to prepare ground-state~\cite{farhi_quantum_2000}. The system's Hamiltonian is then slowly evolved over an annealing time $T$ to the problem Hamiltonian in an adiabatic transition~\cite{born_beweis_1928, kato_adiabatic_1950}
\begin{equation}
    \hat{H}(t) = (1-\nicefrac{t}{T})\hat{H}_B + \nicefrac{t}{T}\hat{H}_P,
\end{equation}
which is a transition where the system stays in its basis state. This process is called quantum annealing and needs to be repeated for multiple measurements, as in a noisy system, not all solutions have the lowest energy. The condition for a sufficiently slow evolution depends mostly on two factors, the temperature of the environment and the spectral gap of the Hamiltonian, i.e. the difference between lowest and second-lowest energy level or eigenvalue. While the first is a system property, the second can be influenced by choosing a suitable Hamiltonian~\cite{benkner_adiabatic_2020}.

The Hamiltonian describing current adiabatic quantum computers such as the \emph{D-wave advantage}, is based on the Ising model~\cite{kadowaki_quantum_1998}. The Ising model uses the Hamiltonian
\begin{equation}
    \hat{H}_{ising} = \sum_{i,j} J_{i,j}\sigma_i\sigma_j + \sum_i h_i \sigma_i,
\end{equation}
where $\sigma \in \{-1, +1\}$ corresponds to the spin of a particle, $J_{i,j}$ represent the interaction between two particles and $h_i$ is an external magnetic field. In an adiabatic quantum computer, the particles' spins are represented by the qubit states and the interactions and external field correspond to the couplings. The lowest energy of the Ising model is equivalent to solving the associated quadratic unconstrained binary optimization (QUBO)
\begin{equation}
    \argmin_\vct{z} \vct{z}^T\vct{Q}\vct{z} + b^T\vct{z},
\end{equation}
which is NP-hard and known to be very challenging for classical solver. As this task can directly be implemented on an adiabatic quantum computer, a considerable speedup for large problem instances is expected in the future.

\section{Quantum MOT}
\label{sec:quantum_mot}
Most existing optimization-based approaches to MOT aim at finding feasible relaxations~\cite{hornakova_making_2021}, implement efficient heuristics in the solution approach~\cite{hornakova_lifted_2020} or use deep learning together with post-processing~\cite{braso_learning_2020} to solve the assignment problem. With the considerable amount of work invested into them, the problem became solvable for growing instances by now. Nevertheless, the assignment problem stays an NP-hard task to solve and growth is thus limited. Quantum computing with the associated speedup on hard problems can provide a solution to this challenge, even if the corresponding optimization problem is much harder to solve with classical approaches at the moment. However, representing tasks in a form suitable for quantum computing often requires a completely new formulation of the problem and MOT is not different in this aspect.

While widely used flow formulations~\cite{li_global_2008, hornakova_lifted_2020, hornakova_making_2021} are suitable for exploiting sparsity, they come with a large set of inequality constraints, which makes them intractable on near-future quantum computers that are limited in the number of qubits. In this context, permutation matrices were shown to be a powerful tool for synchronization or shape matching~\cite{benkner_adiabatic_2020, benkner_q-match_2021, birdal_quantum_2021}. In the following, we therefore propose a formulation based on assignment matrices that grows linearly in the number of required qubits for detections, tracks and frames. Furthermore, it allows to model long-term connections with terms in the cost-matrix that do not require additional qubits.

\paragraph{MOT Formulation.}
\label{sec:track_fromulation}
We approach the MOT problem following the tracking by detection paradigm and use a fixed set of available tracks. Given a set of detections in each frame of a video, appearance features are extracted for each detection. By using a multi-layer Perceptron, pairwise appearance similarities between detections at different timesteps are computed~\cite{hornakova_making_2021}. Starting with this, the goal of the tracking algorithm is to assign each detection to a track, such that the sum of the similarities of detections assigned to a single track is maximized. In this context, a track is defined by its track ID $t$ and each detection in a frame $f$ by its detection ID $d$.

We formulate the given task of assigning detections to a joint set of tracks using assignment matrices, which relax the assumptions of permutation matrices. The binary assignment matrix $\vct{X}_f$ for a frame $f$ maps a vector of detection indices to a vector of tracks at every frame of a video. The elements $x_{dt} \in \{0,1\}$ of the assignment matrix represent the connections between detections $d$ and tracks $t$. Given $D-1$ detections and $T-1$ tracks, the assignment matrix assigns a detection to a track if $x_{dt} = 1$. The requirement that a single detection is assigned to a track at one timestep, leads to the constraint
\begin{equation}
    \label{eq:constr_col_track}
    \sum_{d=1}^D x_{dt} = 1 \quad\forall t \in \{1,...,T-1\}.\\
\end{equation}
And reversely, Equation~\ref{eq:constr_row_det} asserts that every detection is assigned to a single track
\begin{equation}
    \label{eq:constr_row_det}
    \sum_{t=1}^T x_{dt} = 1 \quad\forall d \in \{1,...,D-1\}.\\
\end{equation}

To allow for false-positive detections as well as to handle the case of fewer detections than available tracks, one dummy-detection and one dummy-track, with the respective indices $D$ and $T$, are introduced. A detection assigned to the dummy-track is treated as a false positive and a track that got the dummy-detection assigned to it is inactive or occluded. As the dummy-track and dummy-detection may be assigned multiple times, constraints~\ref{eq:constr_col_track} and~\ref{eq:constr_row_det} do not apply to them.
To model tracks in a sequence consisting of $F$ frames, a single assignment matrix $\vct{X}_{f}$ is required for each frame $f$, mapping the detections to tracks.

\paragraph{Quadratic Form.}
The basis for optimization-based trackers are costs between pairs of detections, where the cost is accounted for if two detections are connected by a common track. The goal of the tracker is to find a solution that minimizes the total cost associated with the assignment. Our approach using assignment matrices leads to a quadratic cost for a pair of frames $i,j$ that reads
\begin{equation}
    \label{eq:cost_explicit}
    c_{ij} = \sum_{t}\sum_{d_i}\sum_{d_j} x_{id_it}q_{d_id_j}x_{jd_jt},
\end{equation}
with $x_{id_it}$ and $x_{jd_jt}$ being entries from the assignment matrices $\vct{X}_i$ and $\vct{X}_j$ respectively and $q_{d_id_j}$ as the corresponding similarity score. It is important to note that only detection pairs assigned to the same track incur a cost, which results in a single sum over the tracks $t$.

Equation~\ref{eq:cost_explicit} can be written in matrix form as
\begin{equation}
    c_{ij} = \text{vec}(\vct{X}_i)^T \vct{Q}_{ij} \text{vec}(\vct{X}_j),
\end{equation}
with $\text{vec}(\vct{X})$ as a row-major vectorization of the corresponding assignment matrices and $\vct{Q}_{ij}$ as the cost matrix of the frame-pair. The maximum frame gap $\Delta f_\text{max}$ that is modeled in our approach depends only on the density of the cost matrix. To include a connection between frames $i$ and $j$, the matrix $\vct{Q}_{ij}$ needs to be filled with the corresponding similarity scores.
The cost matrix $\vct{Q}_{ij}$ is sparse, as it also represents all terms that correspond to detection pairs matched to different tracks, which add no cost. Furthermore, no cost is associated with the mapping of a frame to itself, which includes the main diagonal of $\vct{Q}$.

A complete sequence consisting of $F$ frames, can be represented with the stacked assignment matrix 
\begin{equation}
    \vct{z}=[\text{vec}(\vct{X}_1)^T,..., \text{vec}(\vct{X}_F)^T]^T.
\end{equation}
And the corresponding cost
\begin{equation}
    \label{eq:quad_cost}
    c = \sum_{i=1}^F\sum_{j=1}^F c_{ij} = \vct{z}^T\vct{Q}\vct{z},
\end{equation}
where $\vct{Q}$ is a block-matrix made from all $\vct{Q}_{ij}$.

\paragraph{QUBO form.}
To solve the proposed MOT assignment problem with an adiabatic quantum computer it further needs to be represented as a QUBO task with $\{-1, +1\}$ spin states. This consists of two steps, firstly eliminating the constraints and secondly substituting the variables.

\noindent\textbf{1)} Constraints are represented using a Lagrangian multiplier $\lambda$. As our formulation does not include inequalities, no additional slack variables with corresponding qubits are required. Given the original quadratic program with constraints
\begin{equation}
    \argmin_\vct{z} \vct{z}^T\vct{Q}\vct{z} + \vct{b}^T\vct{z} \quad \text{s.t.} \quad \vct{G}\vct{z} = \vct{d},
\end{equation}
a QUBO can be formulated as
\begin{equation}
    \argmin_\vct{z} \vct{z}^T\vct{Q'}\vct{z} + \vct{b'}^T\vct{z}
\end{equation}
with
\begin{align}
    \label{eq:qubu_penality1}
    \vct{Q'} = \vct{Q} + \lambda\vct{G}^T\vct{G}\\
    \label{eq:qubu_penality2}
    \vct{b'} = -2\lambda\vct{G}^T\vct{b}.
\end{align}

\noindent\textbf{2)} Variables are substituted by replacing the optimization variables $z \in \{0, 1\}$ with $s \in \{-1, 1\}$ by using $z = \nicefrac{1}{2}(s + 1)$ the resulting optimization problem reads
\begin{equation}
    \argmin_\vct{s} \vct{s}^T\vct{Q}\vct{s} + \vct{b}'^T\vct{s} \quad \text{with} \quad \vct{b}'^T = 2(\vct{b}^T + \vct{1}^T\vct{Q}).
\end{equation}

\paragraph{Lagrangian Optimization.} Solving the Lagrangian would require solving a problem in both discrete and continuous optimization variables (assignment, and Lagrangian multipliers, respectively). To solve the problem using AQC, we presented a constant penalty reformulation in the previous paragraph, which fixes the Lagrangian multipliers $\lambda$. In such an approach, if $\lambda$ is large enough, constraint satisfaction is guaranteed. More precisely, a quadratic equality constraint reformulation of the form
\begin{equation}
    \lambda||\vct{G}\vct{z} - \vct{d}||_2^2,
\end{equation}
is used in Equations \ref{eq:qubu_penality1} and \ref{eq:qubu_penality2}, which allows to only consider positive Lagrangian multipliers $\lambda$. Even though $\lambda$ needs to be just large enough from a theoretical perspective, in practice it should be as small as possible. This is especially relevant for AQC, as with a high $\lambda$ the conditioning of the corresponding Hamiltonian in the AQC gets worse. This should be avoided as it results in a lower probability of finding the correct solution in each measurement. 

Thus, in practice a problem dependent bound for the minimum penalty term $\lambda_{\text{min}}$ should be used. One approach to reduce the spectral gap is to estimate an individual $\lambda_i$ for each constraint $\vct{G}_i\vct{x} = \vct{d}_i$ using upper bounds. While such bounds can be computed, they are not tight in many cases. We, therefore, propose a heuristic to estimate the Lagrangian multipliers $\lambda_i$ that closely match their minimal value $\lambda_{i,\text{min}}$. Each multiplier is modeled by
\begin{equation}
    \lambda_i = \lambda_\text{b} + \lambda_i' + \lambda_\text{off},
\end{equation}
where $\lambda_\text{b}$ is a small base value that resolves the easy to fulfill constraints, $\lambda_i'$ is estimated during the optimization procedure and $\lambda_\text{off}$ is an offset to increase the spectral gap.

Starting with $\lambda_i' = 0$ and  $\lambda_\text{off} = 0$ for all constraints, the QUBO is solved using annealing. In general, this will result in a solution $\vct{z}_\lambda$ that does not fulfill the constraints. As in our formulation, only positive violations result in a cost improvement, i.e. $\vct{G}\vct{z} \geq \vct{d}$, the cost reduction of a constraint violation can be estimated as
\begin{align}
    a_i(\vct{z}_\lambda) &= 2(\vct{z}_G^T\vct{Q}\vct{z}_\lambda - \min{(\vct{z}_{G}^T\circ\vct{Q}\vct{z}_\lambda)})/v_i^2, \\
    \vct{z}_G^T &= (\vct{G}_{i}\circ\vct{z}_{\lambda }^{T})\\
    v_i(\vct{z}_\lambda) &= \vct{G}_i\vct{z}_\lambda - \vct{d}_i,
\end{align}
where $\vct{z}_G$ are the variables masked with $\vct{G}_{i}$, $v_i$ is the degree of violation and $\circ$ is the Hadamard product. To fulfill the corresponding constraint, we set
\begin{equation}
    \lambda_i'(\vct{z}_\lambda) = - a_i(\vct{z}_\lambda) - \lambda_\text{b} + \epsilon,
\end{equation}
with a small $\epsilon$ to assert that constraint $i$ is fulfilled in the current setting. While this can be evaluated for all constraints simultaneously, the full procedure needs to be performed iteratively, as not all constraints may be violated in the optimal solution. Nevertheless, the set of measurements returned by the AQC can be used to reduce the number of required iterations. Instead of taking a single best solution, all solutions $\vct{z}_j$ that are close to the optimal solution are evaluated and merged as $\lambda_i' = \max_j \lambda_i'(\vct{z}_j)$. In our formulation, these can be solutions where the track order is permuted.

After estimating the Lagrangian multipliers, the total cost matrix scale is small, nevertheless, the same also holds for the spectral gap, as the cost of not fulfilling constraints is small. Therefore, the additional offset $\lambda_\text{off}$ is added to the Lagrangian multipliers.

\paragraph{Similarity Cost.}
\label{sec:similarity_cost}
We use the same approach for cost generation as AP-lift~\cite{hornakova_making_2021}, where multi-layer Perceptrons are used to regress the similarity score between pairs of detections. Features used to compute this score are the intersection over union (IoU) of aligned boxes and the dot-product between DG-Net~\cite{zheng_joint_2019} appearance features. DG-Net features are generated with the network trained on the MOT15 dataset~\cite{leal-taixe_motchallenge_2015} together with \cite{ristani_performance_2016, wei_person_2018, zheng_scalable_2015}. To generate the MLP input vector, the features are normalized with a global context~\cite{hornakova_lifted_2020}, which results in a total of 22 features~\cite{hornakova_making_2021}. Furthermore, assigning the dummy-detection to a track incurs no cost and assigning a detection to the dummy-track, i.e. labeling it as a false-positive, corresponds to a small negative value $\beta$. This is required to prevent the assignment of single detections to tracks.

\paragraph{Post Processing.}
\label{sec:subtasks}
Even in an offline setting, long sequences cannot be represented as a single optimization problem and need to be split into a set of overlapping subproblems. We set the overlap to the modeled frame gap, and match tracks using the common frames. Matching is stated as a linear sum problem that maximizes the number of detections that are jointly assigned to tracks in both subproblems. As multiple subsequent tracks can be modeled by a single track ID, tracks that are interrupted longer than the maximum modeled frame gap $\Delta f_\text{max}$ are separated.

\paragraph{Problem Scaling.}
One important aspect when designing algorithms for current and near-future quantum computers is the required number of qubits. Many current formulations of MOT grow quickly in size w.r.t. the number of detections, tracks, frames and the length of the modeled frame gap. In contrast to this, the number of qubits in our approach only grows linearly in the number of detections, tracks and frames. Furthermore, by using a quadratic optimization problem, longer frame gaps can be modeled by additional entries in the cost matrix, which correspond to additional couplings between qubits.

While on short sequences the number of possible tracks needs to be at least as high as the total number of tracks, long sequences can profit from a saturation of the required number of tracks. After a track has terminated, there is no cost associated with assigning new detections if they have a distance of more than the maximal frame gap $\Delta f_\text{max}$ from the previous track. Therefore, multiple subsequent real tracks can be modeled by a single track ID and easily be separated in post-processing.

\section{Traditional Solvers}
\label{sec:trad_solver}
While our formulation is advantageous when solved on an adiabatic quantum computer, publicly available real systems have not yet reached a scale where large experiments can be performed. We, therefore, use classical solvers to show the results of our approach on real-world tasks, even though a quadratic problem formulation is known to be hard in this context. A common requirement of solvers to perform quadratic binary optimization via branch and bound is the convexity of the continuous relaxation of the problem. This corresponds to a positive-definite cost matrix $\vct{Q}$, i.e. a matrix with only positive eigenvalues, which is not fulfilled for the given cost matrix in most cases.

\subsection{Hessian Regularization}
\label{sec:pos_def}
A common approach to enforce positive eigenvalues is adding an identity matrix scaled by $\epsilon$. As this changes the cost function and thus the optimal solution, small values need to be used for $\epsilon$, making this approach only suitable for compensating small negative eigenvalues. Nevertheless, investigating the constraints of our formulation leads to a sparse diagonal matrix $\vct{E}$ that can be added to the cost matrix $\vct{Q}$ without changing the optimal solution. With the same approach of grouping the total cost matrix into blocks between frames as in Equation \eqref{eq:quad_cost}, the following definition of $\vct{E}$ is provided in blocks between frames. As only diagonal elements are relevant, blocks between different frames are zero matrices $\vct{E}_{ij} = \vct{0} | i \neq j$. The blocks on the diagonal, which represent the mapping of a frame $i$ to itself $\vct{E}_{ii}$ are diagonal matrices defined by the diagonal elements
\begin{align}
    \label{eq:diag_cost}
    e_{idt} &=
    \begin{cases}
        e & d \in \{1,...,D\} ,t \in \{1,...,T-1\}\\
        0 & t = T\\
    \end{cases}.
\end{align}
The indices refer to the position on the diagonal that correspond to detection $d$ and track $t$. Given a block's assignment matrix $\vct{X}_i$, the total cost of the block after adding the diagonal term is
\begin{equation}
    c_{ii} = \text{vec}(\vct{X}_i)^T (\vct{Q}_{ii} + \vct{E}_{ii}) \text{vec}(\vct{X}_i)
    =e (T-1),
\end{equation}
with $\vct{Q}_{ii} = \vct{0}$ and $T$ tracks in total. The intuition behind the definition is given in the following and the full proof is provided in the supplementary material.

Given a binary problem, any diagonal entry adds cost if a variable is active. In the detection track assignment problem, this corresponds to adding a constant if a detection is assigned to a track. As constraint \ref{eq:constr_col_track} asserts that exactly one detection (real- or dummy-detection) is assigned to every real track each time-step, having a cost $e$ for the assignment adds this cost for each of the $T-1$ real tracks. As the constraint does not apply for the dummy-track with index $T$ and an arbitrary number of detections may be assigned to it. Therefore, the same argument would not hold and we can not add an additional cost to these entries ($e_{ikl}=0|t=T$), without influencing the total cost function.

\section{Experiments and Results}
\label{sec:experiments}
\noindent\textbf{AQC} experiments are performed on a D-wave Advantage 4.1~\cite{mcgeoch_advantage_2021}. The system contains at least 5000 qubits and 35,000 couplers implemented as superconducting qubits~\cite{bunyk_architectural_2014} and Josephson-junctions~\cite{harris_compound_2009} respectively. Every qubit of the D-wave Advantage is connected to 15 other qubits, which needs to be reflected in the sparsity pattern of the cost matrix. If a denser matrix is required, chains of qubits are formed that represent a single state. The actual parameters can vary due to defective qubits and couplers. All experiments are performed using an annealing time of $1600\,\mu \text{s}$ and an additional delay between measurements to reduce the inter-sample correlation. In the following, a single measurement refers to the combination of an annealing cycle and the subsequent measurement.

\noindent\textbf{Simulated annealing} is used to evaluate our approach in a noise-free setting. We use the simulation provided by D-wave for this purpose.

\noindent\textbf{Classical solvers} are used to demonstrate the performance of the proposed algorithm on the full MOT15 dataset. All experiments using classical solvers are performed using Gurobi~\cite{gurobi} with CVXPY~\cite{diamond_cvxpy_2016} as a modeling language.

\subsection{Lagrangian Multiplier}
\label{sec:ex_lagrangian}

\noindent\textbf{Fixed Lagrangian} multipliers represent the basic approach to include constraints in the QUBO. We run experiments with synthetic tracking sequences where object detections are in random order. The scenarios are defined by their similarity scores, which we set to 0.8 for a match and -0.8 for different objects. Furthermore, we add Gaussian noise with variance $\sigma^2$ to the similarity scores and subsequently truncate them to $[-1, 1]$. In the experiments 3 detections over 5 frames and a noise level between $\sigma = 0.2$ to $\sigma = 1.0$ is used. The tracking parameters are set to 4 tracks and a maximal frame-gap of $\Delta f_\text{max} = 3$ frames.

\begin{figure}
     \centering
     \begin{subfigure}[b]{\linewidth}
        \centering
        \includegraphics[width=\linewidth, trim= -15 39 -18 0, clip]{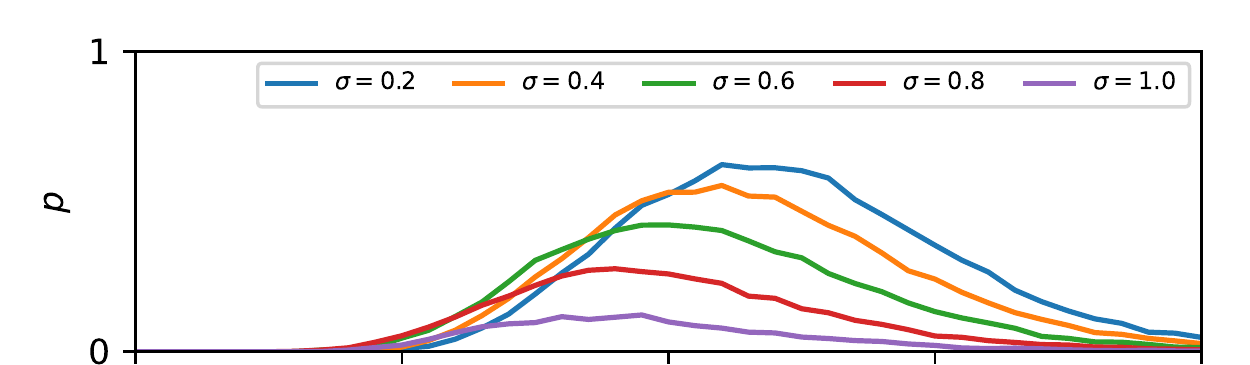}
        \label{fig:lagrange_sim_summary}
     \end{subfigure}
     \hfill
     \vspace{-20pt}
     \begin{subfigure}[b]{\linewidth}
         \centering
         \includegraphics[width=\textwidth, trim= 0 12 -11 10, clip]{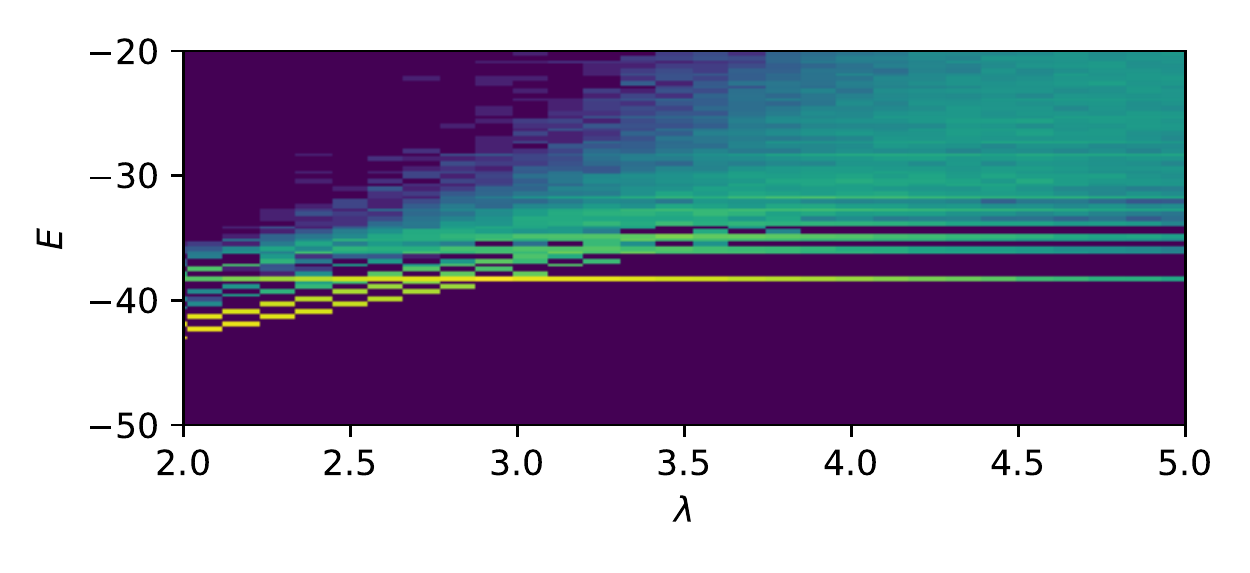}
         \label{fig:lagrange_sim_n0_2}
     \end{subfigure}
    \vspace{-25pt}
    \caption{Solution probability and energy levels using simulated annealing for different noise levels and changing $\lambda$.}
    \label{fig:lagrange_sim}
\end{figure}

Results generated with simulated annealing are shown in Figure~\ref{fig:lagrange_sim}, where the top plot shows the solution probability for different noise levels over an increasing Lagrangian multiplier. For each $\lambda$, $4096$ measurements are performed. The lower plot shows the histogram over the energy of the returned solutions for a noise level of $\sigma=0.6$. The correct solution can be seen at an energy level of $-38.6$.

With increasing noise level, the solution probability for the best value of $\lambda$ reduces considerably, which can be explained by the energy histogram. As described in Section \ref{sec:aqc}, a low spectral gap, i.e. the difference between the lowest and second-lowest energy level, reduces the probability of the AQC staying in its ground state and thus, the probability of finding the correct solution. In the energy plot, the spectral gap is visible as the distance between the energy band of the correct solution and the next higher energy band, given a sufficiently high $\lambda$, such that the correct solution has the lowest energy.

\begin{figure}
     \centering
     \begin{subfigure}[b]{\linewidth}
        \centering
        \includegraphics[width=\linewidth, trim=  1 39 -11 0, clip]{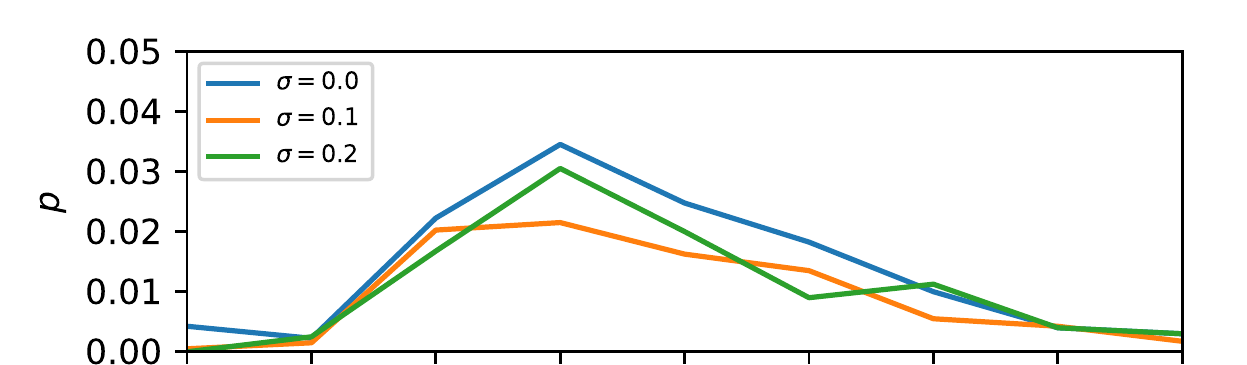}
        \label{fig:lagrange_real_summary}
     \end{subfigure}
     \hfill
     \vspace{-20pt}
     \begin{subfigure}[b]{\linewidth}
         \centering
         \includegraphics[width=\textwidth, trim= 0 12 -10 10, clip]{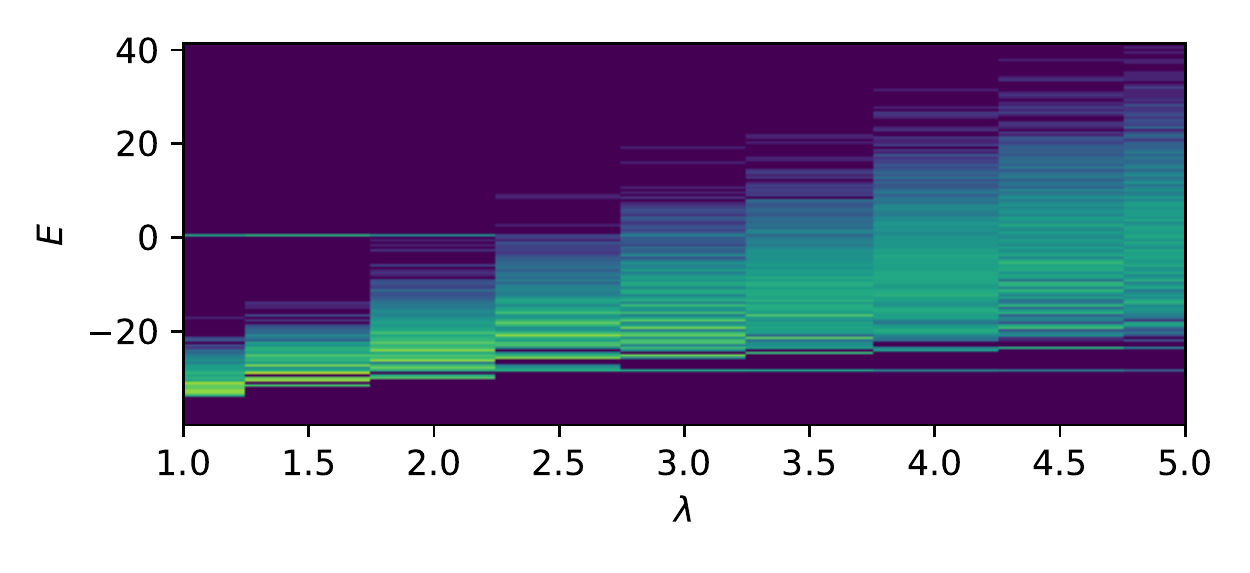}
         \label{fig:lagrange_real_n0_0}
     \end{subfigure}
    \vspace{-25pt}
    \caption{Solution probability and energy levels using quantum annealing for different noise levels and changing $\lambda$.}
    \label{fig:lagrange_real}
\end{figure}

Tracking with the D-wave advantage is performed on a problem with 3 detections over 4 frames and noise levels $\sigma \in \{0.0, 0.1, 0.2\}$. Results using 4000 measurements for each setting are shown in Figure~\ref{fig:lagrange_real}. Solution probabilities are lower compared to simulated annealing and high energy solutions are returned more often. This can be explained by the high noise of current AQCs.

\begin{figure}
     \centering
     \begin{subfigure}[b]{\linewidth}
        \centering
        \includegraphics[width=\linewidth, trim= -15 39 0 0, clip]{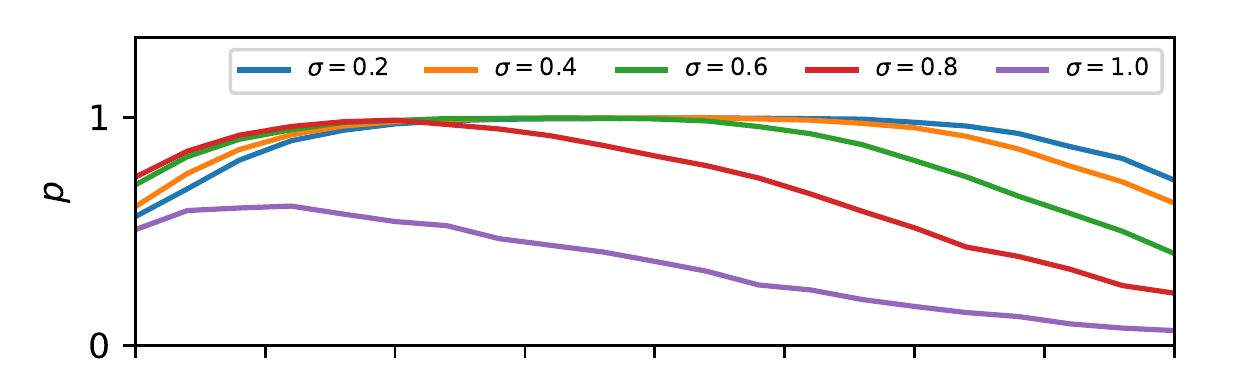}
        \label{fig:lagrange_off_sim_summary}
     \end{subfigure}
     \hfill
     \vspace{-20pt}
     \begin{subfigure}[b]{\linewidth}
         \centering
         \includegraphics[width=\textwidth, trim= 0 12 0 10, clip]{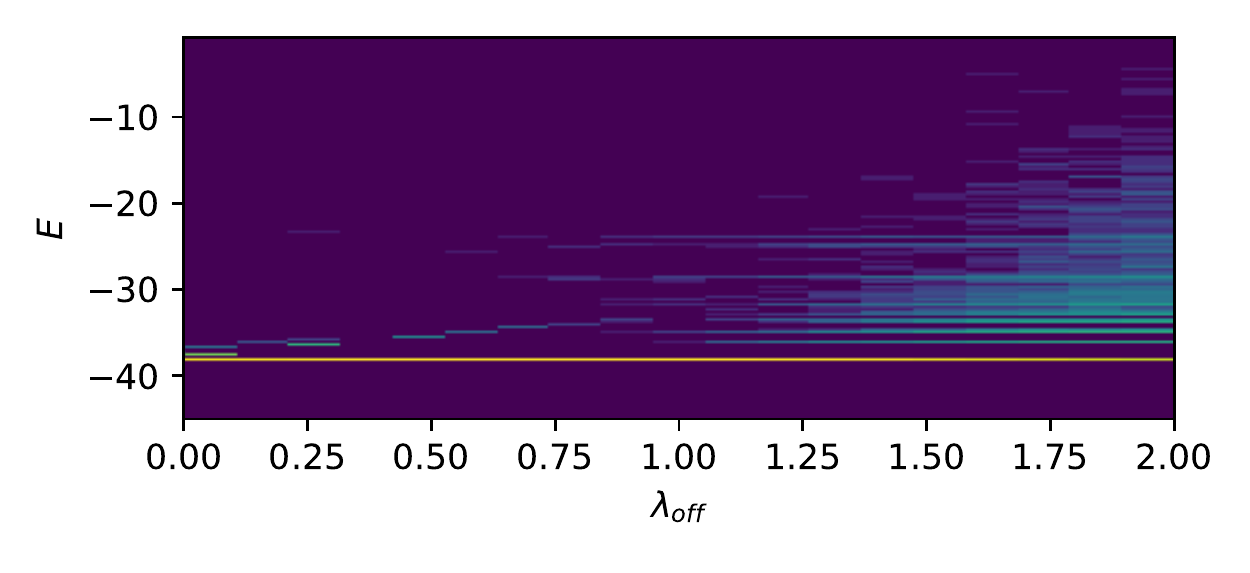}
         \label{fig:lagrange_off_sim_n0_2}
     \end{subfigure}
    \vspace{-25pt}
    \caption{Solution probability and energy levels using simulated annealing and optimized $\lambda_i$ for different noise levels over $\lambda_\text{off}$.}
    \label{fig:lagrange_off_sim}
\end{figure}

\begin{figure}
     \centering
     \begin{subfigure}[b]{\linewidth}
        \centering
        \includegraphics[width=\linewidth, trim= 1 39 0 0, clip]{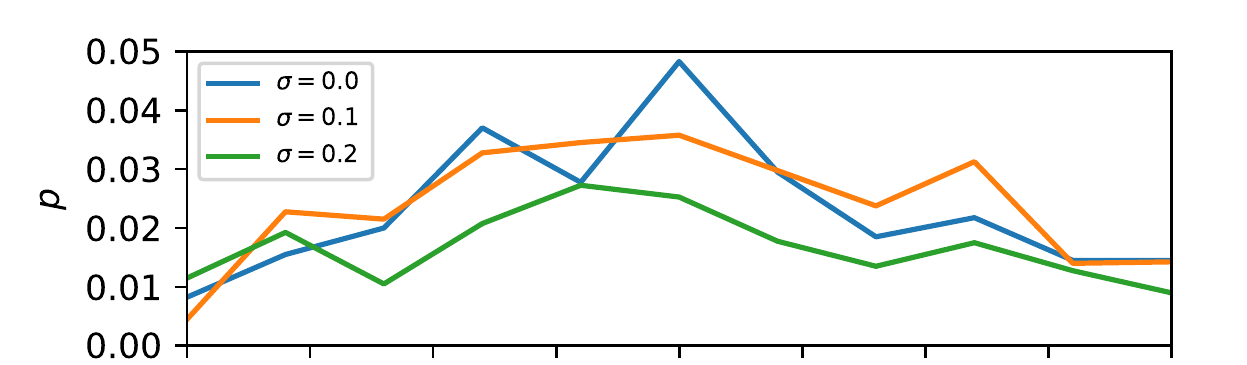}
        \label{fig:lagrange_off_real_summary}
     \end{subfigure}
     \hfill
     \vspace{-20pt}
     \begin{subfigure}[b]{\linewidth}
         \centering
         \includegraphics[width=\textwidth, trim= 0 12 -12 10, clip]{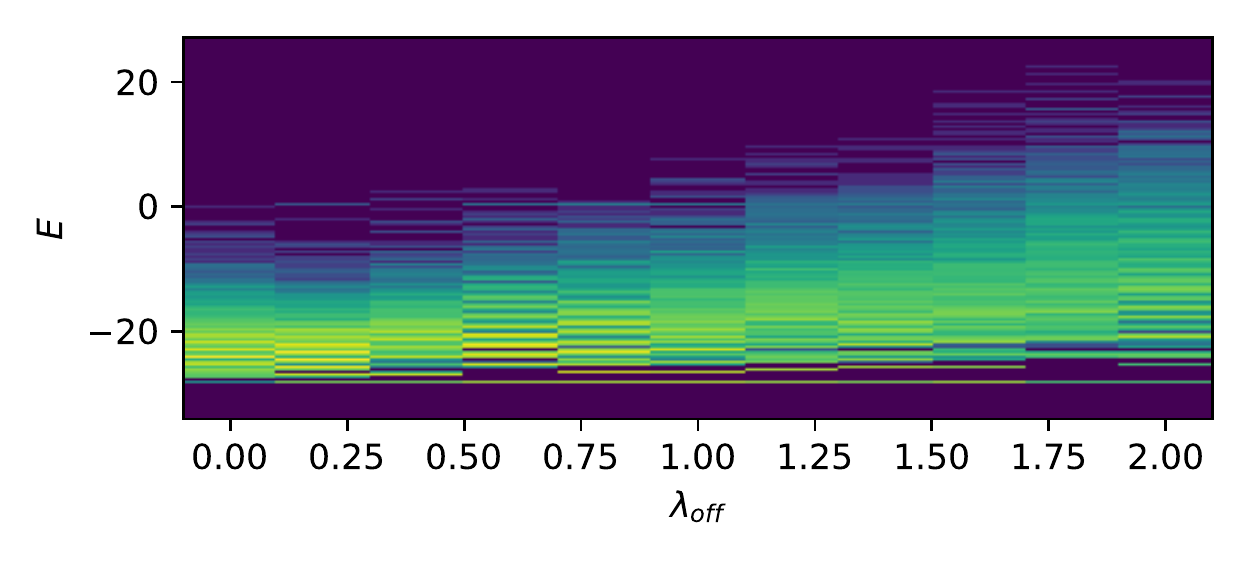}
         \label{fig:lagrange_off_real_n0_0}
     \end{subfigure}
    \vspace{-25pt}
    \caption{Solution probability and energy levels using quantum annealing and optimized $\lambda_i$ for different noise levels over $\lambda_\text{off}$.}
    \label{fig:lagrange_off_real}
\end{figure}

\paragraph{Optimized Lagrangian} multipliers are introduced to improve the spectral gap of the normalized cost matrix. We perform the same tracking tasks as for fixed Lagrangian multipliers, but evaluate the results w.r.t. the offset term $\lambda_\text{off}$. Results generated with simulated annealing are shown in Figure~\ref{fig:lagrange_off_sim}. Optimization of the Lagrangian multipliers is initialized with a base value of $\lambda_b = 0.5$. The probability of finding the right solution is increased and stays high over a large range of $\lambda_\text{off}$ compared to only using a single $\lambda$. Furthermore, the best solution probability for each of the noise levels is better than the optimum for a fixed Lagrangian multiplier. This has two advantages: first, fewer measurements are needed to find the correct solution and secondly, less effort needs to be invested to find a good setting for $\lambda$. Results for the problem with an optimized Lagrangian multiplier with $\lambda_b=1.0$ solved on the AQC are shown in Figure~\ref{fig:lagrange_off_real}. When optimally tuned for $\sigma = 0$, our method returns the best solution in $4.8\%$ of the measurements, compared to $3.5\%$ when using a fixed multiplier. Furthermore, even without an additional offset $\lambda_\text{off} = 0$, the best solution is returned in $0.8\%$ of the measurements.

\subsection{MOT15}
\label{sec:mot15}
We use the MOT15 dataset~\cite{leal-taixe_motchallenge_2015} to show that our method performs on par with state-of-the-art tracking methods. For this dataset, GUROBI~\cite{gurobi} is used to find a solution for the optimization problem. The sequence is evaluated in segments of 20 frames using a maximum frame gap of $\Delta f_\text{max} = 10$. As binary quadratic problems are very hard to solve with classical approaches, it is not possible to find an optimum solution for segments that contain a high number of tracks. In these cases, we terminate the optimization after $900\,\text{s}$ on a single segment and use the best solution found.

For comparisons, ApLift~\cite{hornakova_making_2021} is closest to our method, as it uses the same set of similarity features. On the test set, we achieve a MOTA-score of $49.9\%$ and perform only $1.2\%$ below ApLift, even though it models gaps up to $50$ frames.

For a comparison under similar settings, we evaluate our method and ApLift~\cite{hornakova_making_2021} with the same frame gap of $\Delta f_\text{max} = 10$. As MOT15 does not contain a validation set, we use leave one out cross-validation on all samples of the training set for a fair comparison. In this scenario, our method improves by $0.2\%$ over ApLift in $MOTA$ score. An explanation for this is that the MOT15 test set contains more detections in each frame on average (10.6 vs. 7.3) than the training set. In this case, there are more sequences where the classical solver does not find a solution and thus, generates a non-optimal result.

\begin{figure}
     \centering
     \begin{subfigure}[b]{0.24\linewidth}
        \centering
        \includegraphics[width=\linewidth, trim= 400 200 0 100, clip]{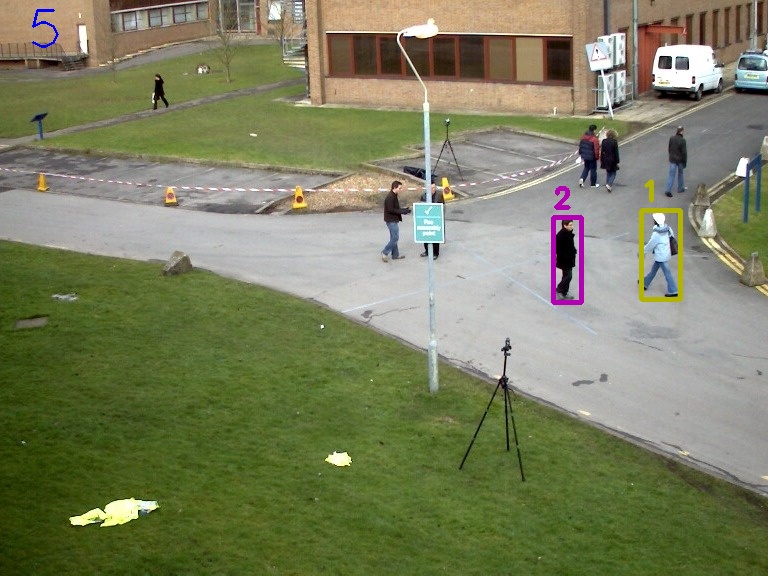}
     \end{subfigure}
          \begin{subfigure}[b]{0.24\linewidth}
        \centering
        \includegraphics[width=\linewidth, trim= 400 200 0 100, clip]{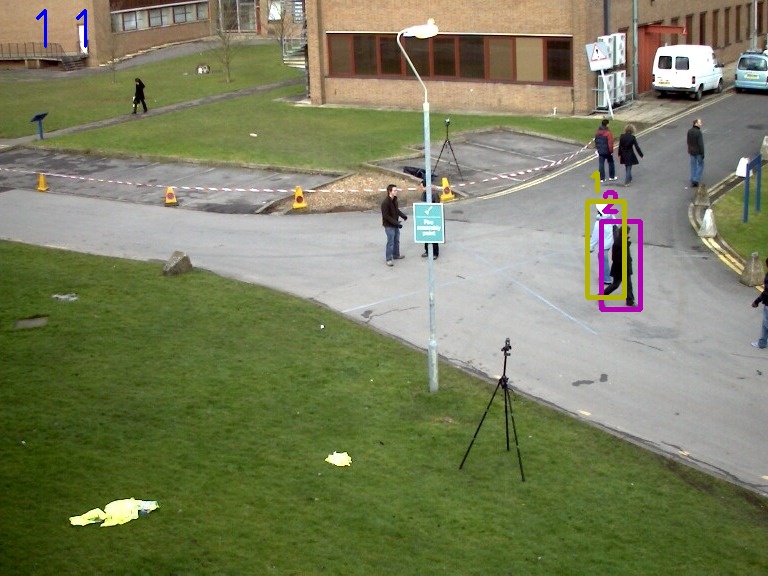}
     \end{subfigure}
          \begin{subfigure}[b]{0.24\linewidth}
        \centering
        \includegraphics[width=\linewidth, trim= 400 200 0 100, clip]{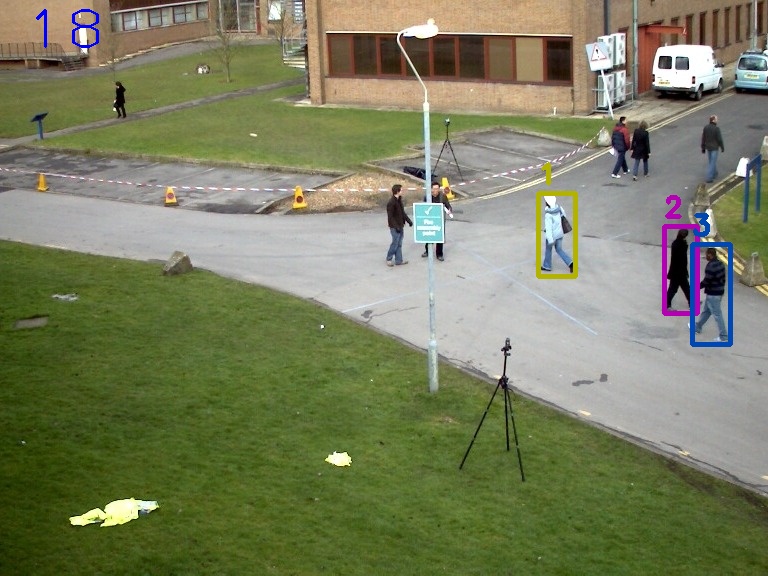}
     \end{subfigure}
          \begin{subfigure}[b]{0.24\linewidth}
        \centering
        \includegraphics[width=\linewidth, trim= 400 200 0 100, clip]{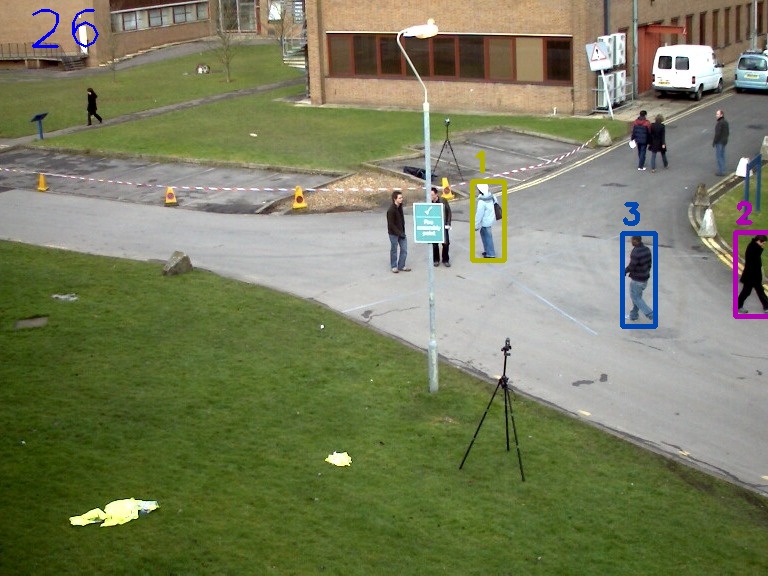}
     \end{subfigure}
    \caption{Frames from the extracted sequence tracked on the AQC.}
    \label{fig:mot15_real_visualize}
\end{figure}

\begin{figure}
\centering
\includegraphics[width=\linewidth, trim= 10 5 38 20, clip]{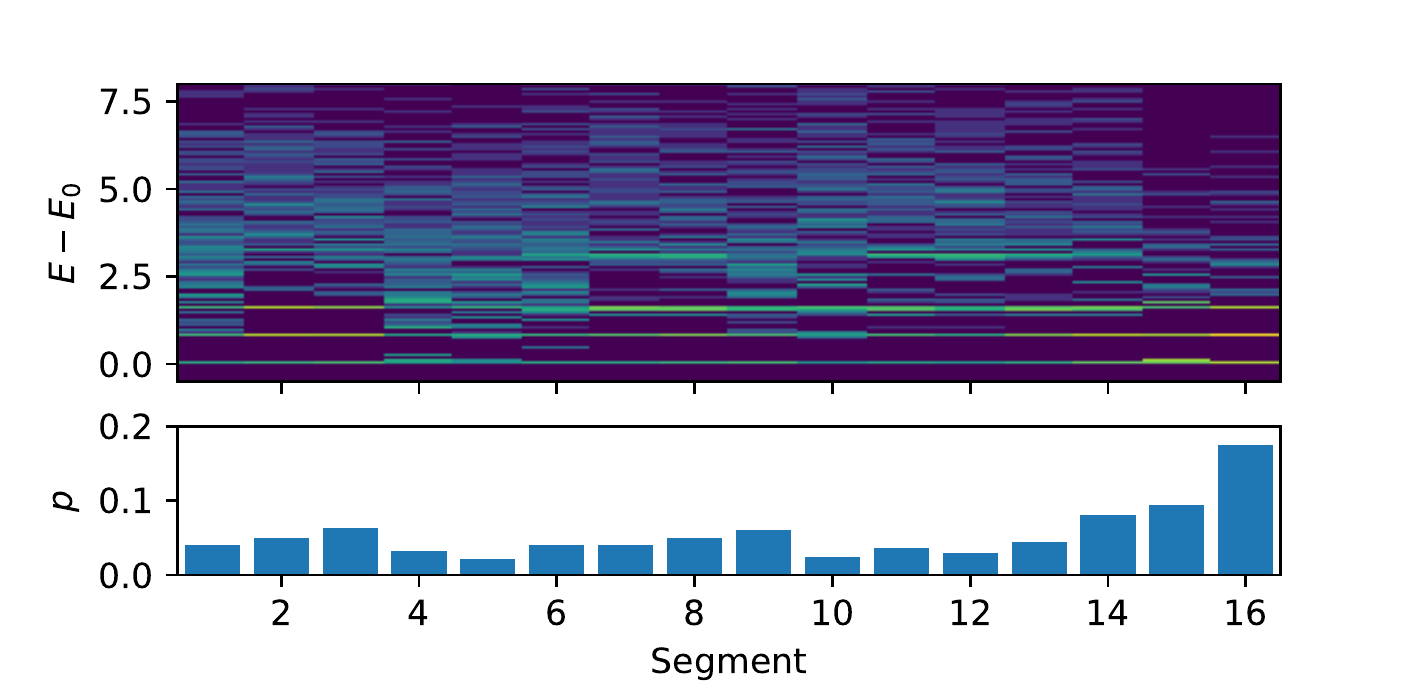}
\caption{Energy of measurements returned by performing tracking of the PETS09-S2L1 sequence on the D-wave Advantage. The bar-plot shows the probability of measuring the optimal solution.}
\label{fig:mot15_aqc}
\end{figure}

\begin{table}[tb]
\vspace{3pt}
\begin{center}
\small
\input{tables/mot15_short}
\end{center}
\vspace{-10pt}
\caption{Results on MOT15~\cite{leal-taixe_motchallenge_2015}. X-val refers to results on the training set using leave-one-out cross validation.}
\label{tab:mot15}
\end{table}
\vspace{2pt}
\noindent\textbf{MOT15 with AQC.}
\label{sec:mot15_aqc}
To show that tracking with an AQC already scales to small real-world examples, a part of the PETS09-S2L1 sequence is used. As the problem size has to be limited, three tracks that contain two occlusions, are extracted between frames 121 to 155. We execute our pipeline on segments of 5 frames with 3 tracks, a maximum frame-gap of 3, and optimized Lagrangian multipliers. The subproblems are solved on the D-wave Advantage with $1600\,\mu \text{s}$ annealing time and 500 measurements per segment. The most relevant frames that highlight occlusions are shown in Figure~\ref{fig:mot15_real_visualize}. The normalized energy $E-E_0$ levels of the measurements for each subproblem are shown at the top of Figure~\ref{fig:mot15_aqc} and the corresponding probabilities $p$ of measuring the right solutions are plotted in the lower one. The subproblems 5 and 10 correspond to the two occlusions highlighted in Figure~\ref{fig:mot15_real_visualize}. These are harder to solve problems, as multiple solutions with small differences in their energy exist and thus, they have a lower solution probability.

\section{Conclusion}
In this work, we proposed the first quantum computing formulation of MOT. We demonstrated that current AQCs can solve small real-world tracking problems, and that our approach closely matches state-of-the-art MOT methods. Current limitations stem from the proposed formulation being optimized to run on an AQC. As QUBO is know to be hard using classical approaches and as current AQCs are still at an experimental stage, problems are limited to a small scale. Nevertheless, quantum computing has the potential to make much larger problems feasible in the future.


\input{references.bbl}
\clearpage

\begin{center}
\textbf{\large Supplementary: Adiabatic Quantum Computing for Multi Object Tracking}
\end{center}

\section{Introduction}
\label{s_sec:intro}
The supplementary material aims at giving a more thorough insight into the technical details of our work and at highlighting results obtained using simulated and quantum annealing. First, we prove that Hessian regularization does not influence the minimizer of the binary optimization problem in Section~\ref{s_sec:pos_def_detail}. After this, more details on matching multiple subproblems in post processing are provided in Section~\ref{s_sec:post_processing} and a further analysis of measurements generated using simulated and quantum annealing is presented in Section~\ref{s_sec:annealing_energy}. Finally, detailed results on the MOT15 challenge are shown in Section~\ref{s_sec:mot15_details} and furthermore, also visualized in the accompanying video.

\section{Hessian Regularization}
\label{s_sec:pos_def_detail}
The following proof shows that the optimum solution is not influenced by the additional diagonal terms introduced in Section~5.1 of the main paper. This holds given a binary optimization problem and the constraints in Equations~(6) and (7).

\begin{align}
    c_{ii} &= \text{vct}(\vct{X}_i^T)\vct{E}_{ii}\text{vct}(\vct{X}_i)\\
    &=^\text{diag}  \sum_{t=1}^T \sum_{d=1}^D x_{id_it}^2 e_{d_it}\\
    &=^{\text{bin}} \sum_{t=1}^T \sum_{d=1}^D x_{id_it} e_{d_it}\\
    &= \sum_{t=1}^{T-1} \sum_{d=1}^D x_{id_it} e_{d_it} + \sum_{d=1}^D x_{id_iT} e_{d_iT}\\
    &=^{(22)} \sum_{t=1}^{T-1} \sum_{d=1}^D x_{id_it} e + \sum_{d=1}^D x_{id_iT} 0\\
    &= e \sum_{t=1}^{T-1} \sum_{d=1}^D x_{id_it}\\
    &=^{(6)} e \sum_{t=1}^{T-1} 1\\
    &= e (T-1)
\end{align}

\section{Post Processing}
\label{s_sec:post_processing}
To allow the handling of long sequences that cannot be represented as a single optimization problem, the sequence needs to be split into overlapping subproblems. We split a long sequence in equally sized subproblems with an overlap similar to the modeled frame gap. After tracking each subproblem separately, tracks are matched between each pair of neighboring subproblems by solving a linear sum problem that can be solved in polynomial time. The optimization goal is to maximize the number of detections that are jointly assigned to tracks matched in both subproblems. The linear sum optimization problem for matching subproblems $k$ and $k+1$ is stated as
\begin{equation}
    \renewcommand{\arraystretch}{1.5}
   \max_{{x_{ij} \in \{0,1\}}}  \sum_{i=1}^{T_k} \sum_{j=1}^{T_{k+1}} x_{ij}\,m_{ij} \quad \text{s.t.} \quad  \begin{array}{l}
        \sum_{i=1}^{T_k}  \hspace{1.6mm} x_{ij} \leq 1 \\
        \sum_{j=1}^{T_{k+1}} x_{ij} \leq 1,
    \end{array} 
\end{equation}
where $x_{ij}$ are the optimization variables indicating an assignment of track $i$ in segment $k$ to track $j$ in segment $k+1$, The considered tracks $T_k$ and $T_{k+1}$ are the tracks that have at least one detection assigned to them in the frames overlapping between both subproblems. $m_{ij}$ is the number of detections shared by tracks $i$ and $j$ in the overlapping frames, which furthermore is set to a small negative value if tracks $i$ and $j$ have no overlap.

\section{Annealing Energy Distribution}
\label{s_sec:annealing_energy}
\begin{figure}
     \centering
     \begin{subfigure}[b]{\linewidth}
        \centering
        \includegraphics[width=\linewidth, trim= -15 39 -18 0, clip]{figures/lagrange/lagrange_results_accumulated.pdf}
        \label{s_fig:lagrange_sim_summary}
     \end{subfigure}
     \hfill
     \vspace{-20pt}
     \begin{subfigure}[b]{\linewidth}
         \centering
         \includegraphics[width=\textwidth, trim= 0 39 -11 10, clip]{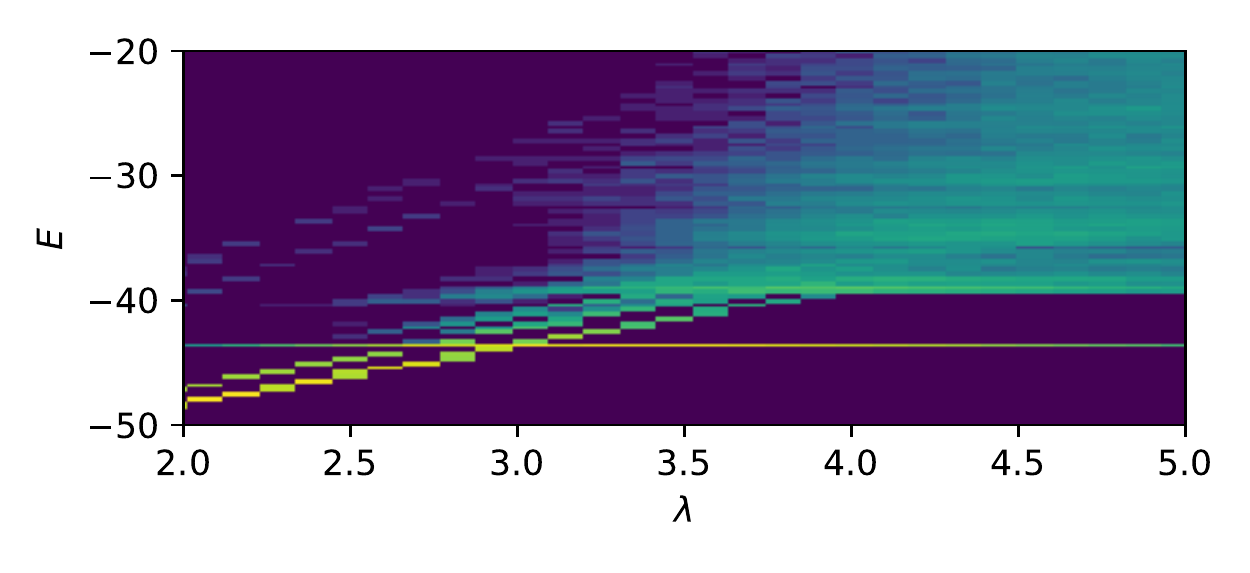}
         \label{s_fig:lagrange_sim_n0_2}
     \end{subfigure}
     \hfill
     \vspace{-20pt}
     \begin{subfigure}[b]{\linewidth}
         \centering
         \includegraphics[width=\textwidth, trim= 0 39 -11 10, clip]{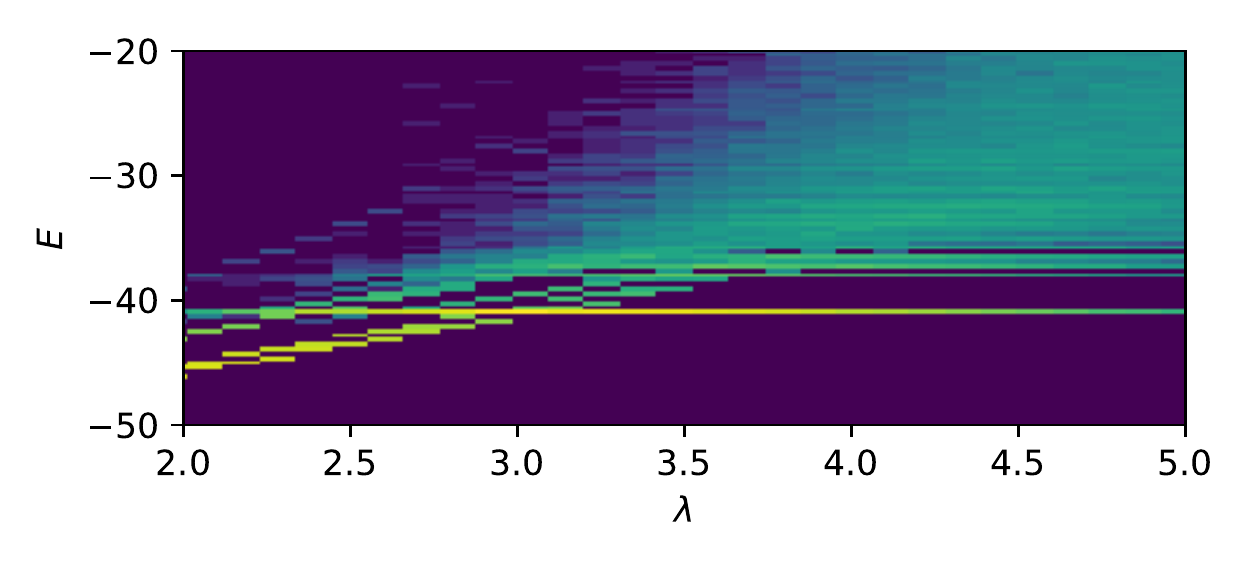}
         \label{s_fig:lagrange_sim_n0_4}
     \end{subfigure}
     \hfill
     \vspace{-20pt}
     \begin{subfigure}[b]{\linewidth}
         \centering
         \includegraphics[width=\textwidth, trim= 0 39 -11 10, clip]{figures/lagrange/lagrange_results_noise_0_6.pdf}
         \label{s_fig:lagrange_sim_n0_6}
     \end{subfigure}
     \hfill
     \vspace{-20pt}
     \begin{subfigure}[b]{\linewidth}
         \centering
         \includegraphics[width=\textwidth, trim= 0 39 -11 10, clip]{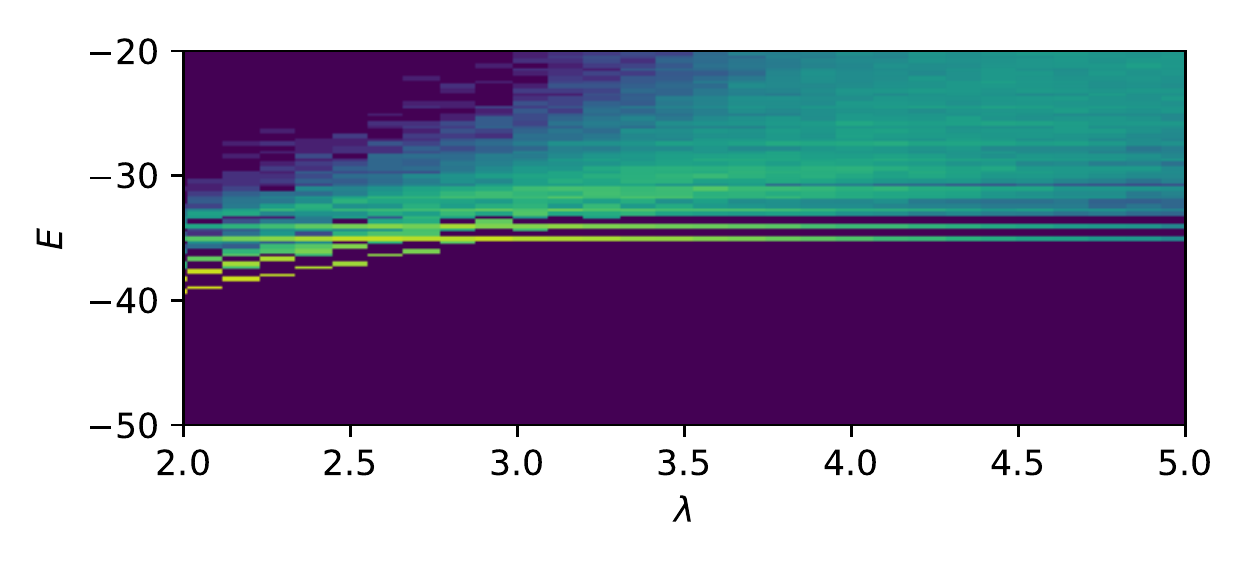}
         \label{s_fig:lagrange_sim_n0_8}
     \end{subfigure}
     \hfill
     \vspace{-20pt}
     \begin{subfigure}[b]{\linewidth}
         \centering
         \includegraphics[width=\textwidth, trim= 0 12 -11 10, clip]{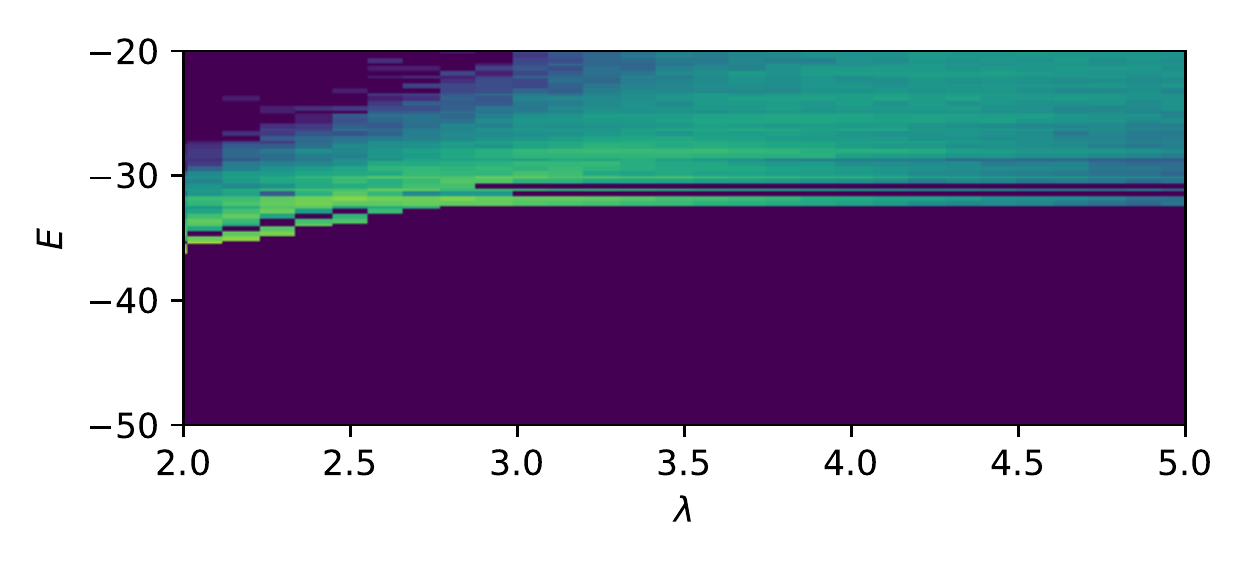}
         \label{s_fig:lagrange_sim_n1_0}
     \end{subfigure}
    \vspace{-25pt}
    \caption{Solution probability and energy levels using simulated annealing for noise levels $\sigma \in \{0.2,0.4,0.6,0.8,1.0\}$ and changing $\lambda$.}
    \label{s_fig:lagrange_sim}
\end{figure}

\begin{figure}
     \centering
     \begin{subfigure}[b]{\linewidth}
        \centering
        \includegraphics[width=\linewidth, trim=  1 39 -11 0, clip]{figures/lagrange_real/lagrange_results_accumulated.pdf}
        \label{s_fig:lagrange_real_summary}
     \end{subfigure}
     \hfill
     \vspace{-20pt}
     \begin{subfigure}[b]{\linewidth}
         \centering
         \includegraphics[width=\textwidth, trim= 0 39 -10 10, clip]{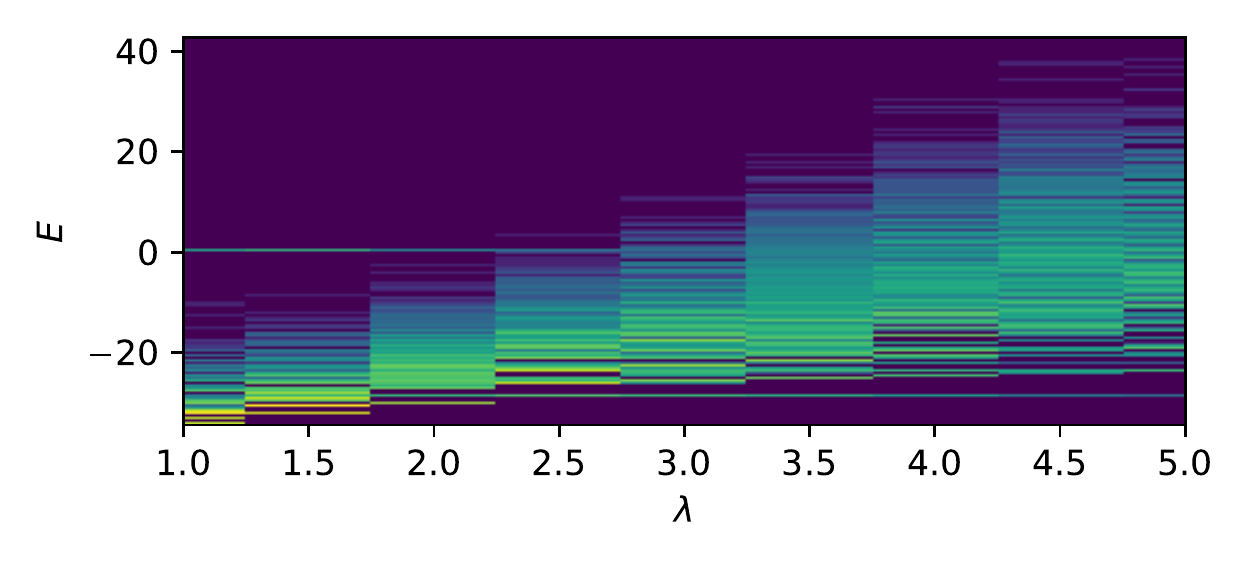}
         \label{s_fig:lagrange_real_n0_0}
     \end{subfigure}
     \hfill
     \vspace{-20pt}
     \begin{subfigure}[b]{\linewidth}
         \centering
         \includegraphics[width=\textwidth, trim= 0 39 -10 10, clip]{figures/lagrange_real/lagrange_results_noise_0_1.pdf}
         \label{s_fig:lagrange_real_n0_1}
     \end{subfigure}
     \hfill
     \vspace{-20pt}
     \begin{subfigure}[b]{\linewidth}
         \centering
         \includegraphics[width=\textwidth, trim= 0 12 -10 10, clip]{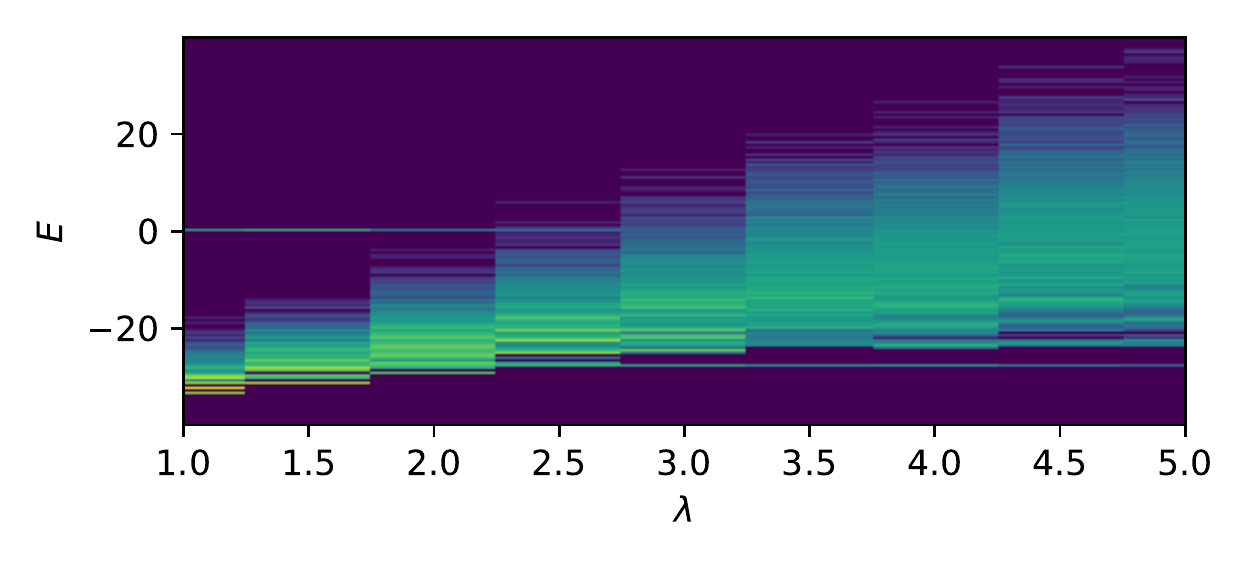}
         \label{s_fig:lagrange_real_n0_2}
     \end{subfigure}
    \vspace{-25pt}
    \caption{Solution probability and energy levels using quantum annealing for noise levels $\sigma \in \{0.0,0.1,0.2\}$ over $\lambda$.}
    \label{s_fig:lagrange_real}
\end{figure}

\begin{figure}
     \centering
     \begin{subfigure}[b]{\linewidth}
        \centering
        \includegraphics[width=\linewidth, trim= -15 39 0 0, clip]{figures/lagrange_offset/lagrange_offset_results_accumulated.pdf}
        \label{s_fig:lagrange_off_sim_summary}
     \end{subfigure}
     \hfill
     \vspace{-20pt}
     \begin{subfigure}[b]{\linewidth}
         \centering
         \includegraphics[width=\textwidth, trim= 0 39 0 10, clip]{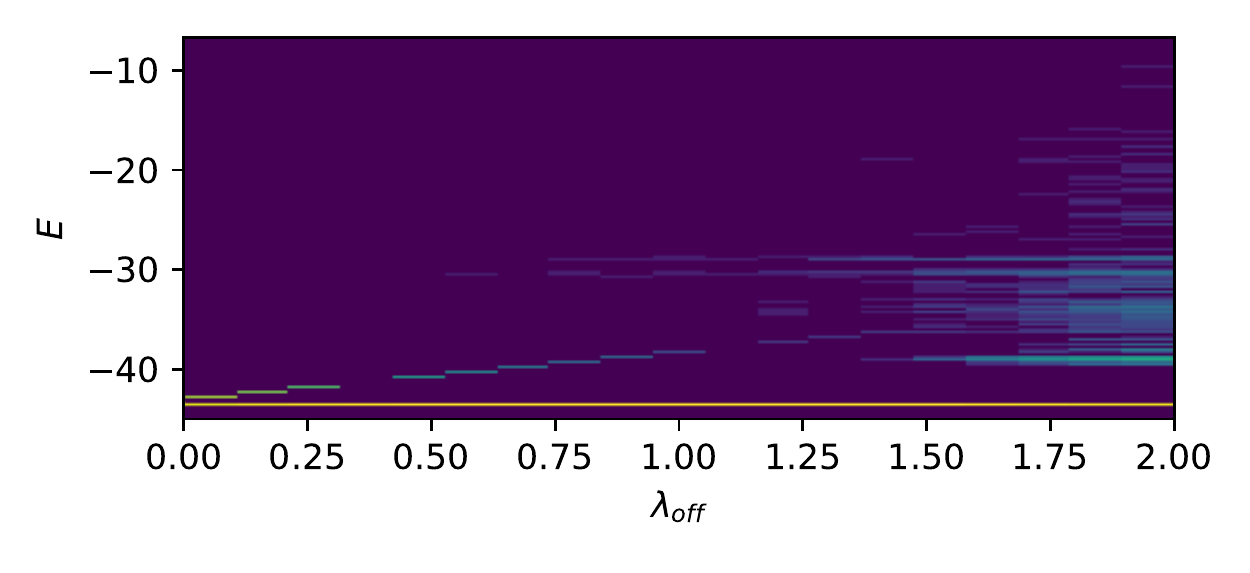}
         \label{s_fig:lagrange_off_sim_n0_2}
     \end{subfigure}
     \hfill
     \vspace{-20pt}
     \begin{subfigure}[b]{\linewidth}
         \centering
         \includegraphics[width=\textwidth, trim= 0 39 0 10, clip]{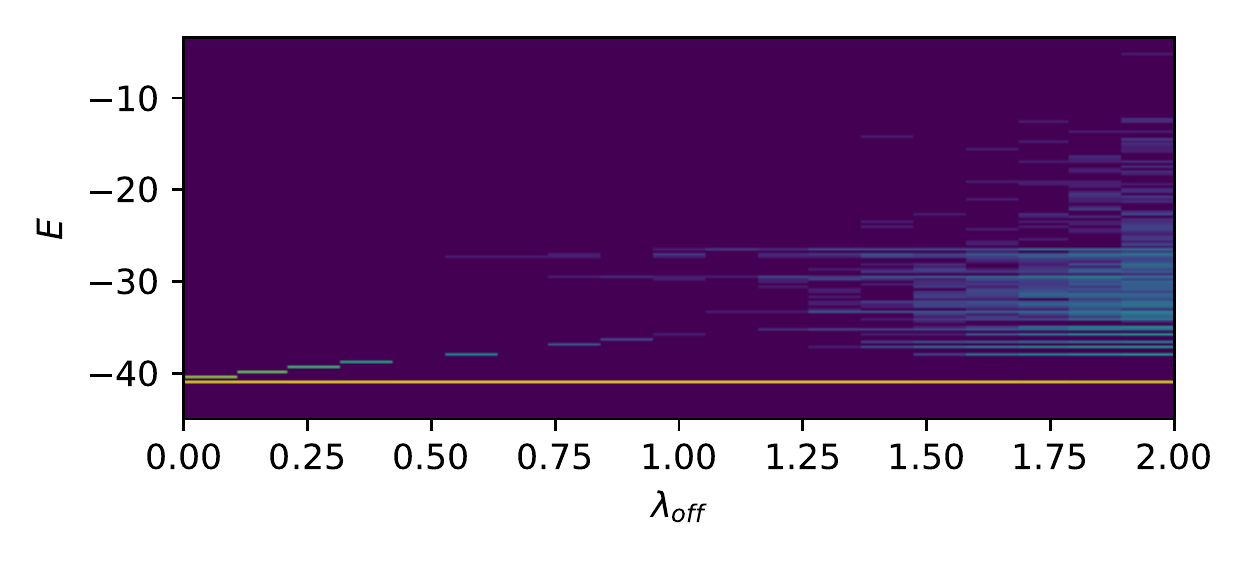}
         \label{s_fig:lagrange_off_sim_n0_4}
     \end{subfigure}
     \hfill
     \vspace{-20pt}
     \begin{subfigure}[b]{\linewidth}
         \centering
         \includegraphics[width=\textwidth, trim= 0 39 0 10, clip]{figures/lagrange_offset/lagrange_offset_results_noise_0_6.pdf}
         \label{s_fig:lagrange_off_sim_n0_6}
     \end{subfigure}
     \hfill
     \vspace{-20pt}
     \begin{subfigure}[b]{\linewidth}
         \centering
         \includegraphics[width=\textwidth, trim= 0 39 0 10, clip]{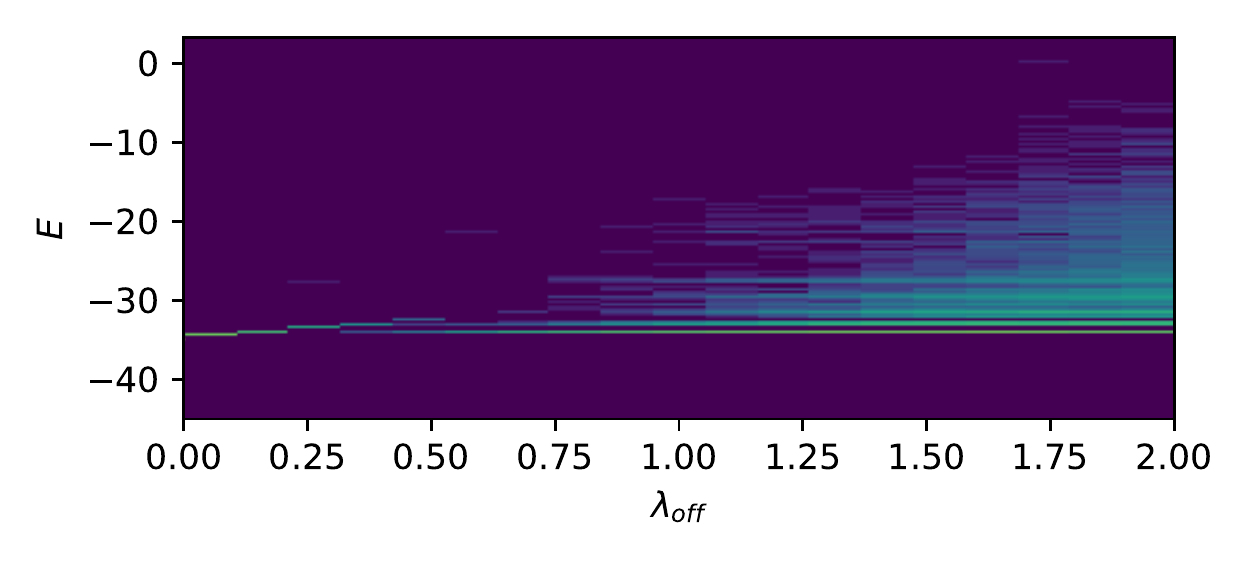}
         \label{s_fig:lagrange_off_sim_n0_8}
     \end{subfigure}
     \hfill
     \vspace{-20pt}
     \begin{subfigure}[b]{\linewidth}
         \centering
         \includegraphics[width=\textwidth, trim= 0 12 0 10, clip]{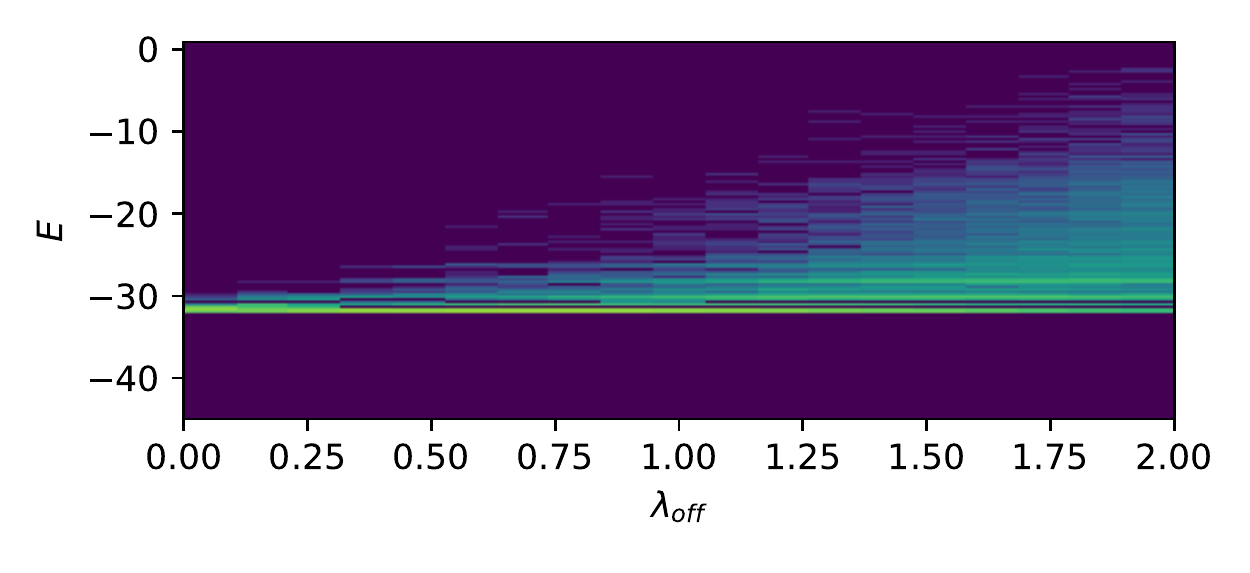}
         \label{s_fig:lagrange_off_sim_n1_0}
     \end{subfigure}
    \vspace{-25pt}
    \caption{Solution probability and energy levels using simulated annealing and optimized $\lambda_i$ for noise levels $\sigma \in \{0.2,0.4,0.6,0.8,1.0\}$ over $\lambda_\text{off}$.}
    \label{s_fig:lagrange_off_sim}
\end{figure}

\begin{figure}
     \centering
     \begin{subfigure}[b]{\linewidth}
        \centering
        \includegraphics[width=\linewidth, trim= 1 39 0 0, clip]{figures/lagrange_offset_real/lagrange_offset_results_accumulated.pdf}
        \label{s_fig:lagrange_off_real_summary}
     \end{subfigure}
     \hfill
     \vspace{-20pt}
     \begin{subfigure}[b]{\linewidth}
         \centering
         \includegraphics[width=\textwidth, trim= 0 39 -12 10, clip]{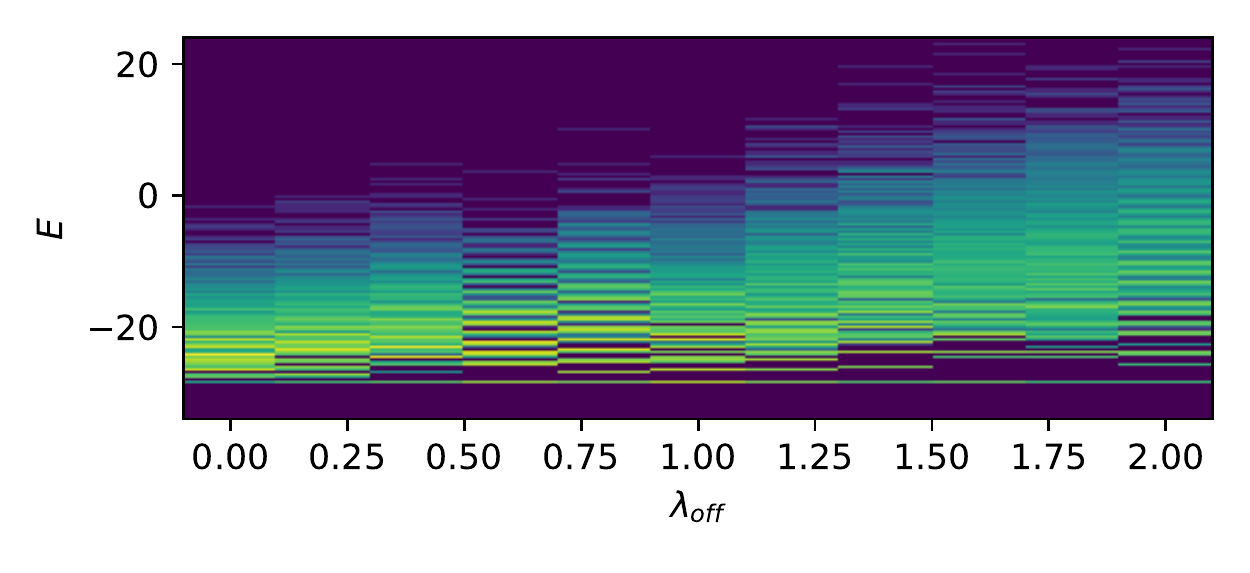}
         \label{s_fig:lagrange_off_real_n0_0}
     \end{subfigure}
     \hfill
     \vspace{-20pt}
     \begin{subfigure}[b]{\linewidth}
         \centering
         \includegraphics[width=\textwidth, trim= 0 39 -12 10, clip]{figures/lagrange_offset_real/lagrange_offset_results_noise_0_1.pdf}
         \label{s_fig:lagrange_off_real_n0_1}
     \end{subfigure}
     \hfill
     \vspace{-20pt}
     \begin{subfigure}[b]{\linewidth}
         \centering
         \includegraphics[width=\textwidth, trim= 0 12 -12 10, clip]{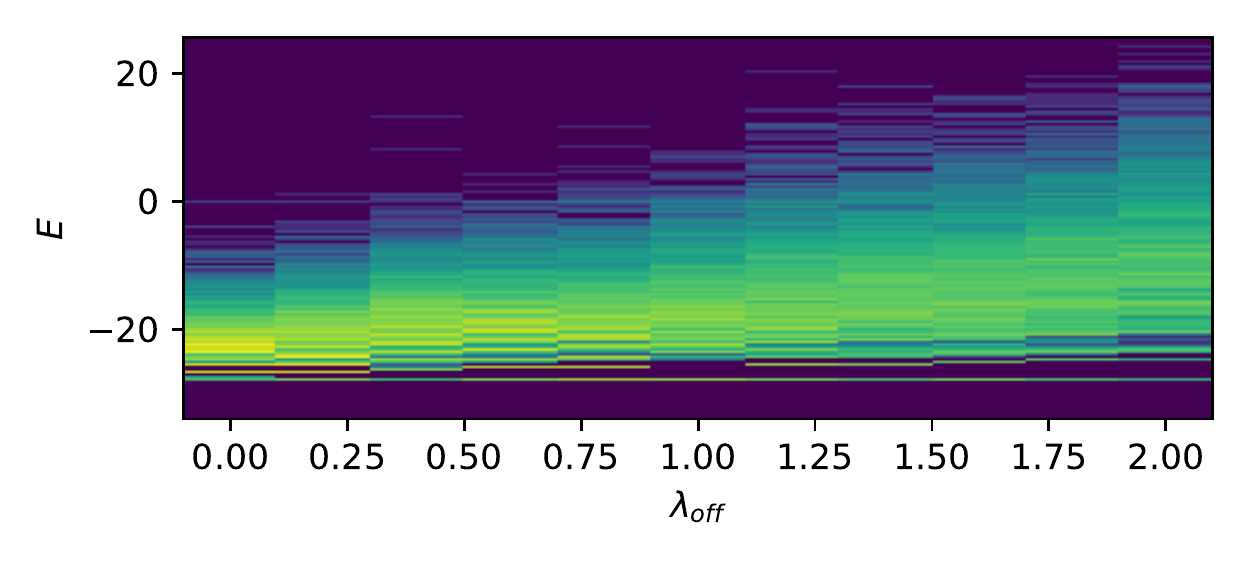}
         \label{s_fig:lagrange_off_real_n0_2}
     \end{subfigure}
    \vspace{-25pt}
    \caption{Solution probability and energy levels using quantum annealing and optimized $\lambda_i$ for noise levels $\sigma \in \{0.0,0.1,0.2\}$ over $\lambda_\text{off}$.}
    \label{s_fig:lagrange_off_real}
\end{figure}

Results from simulated as well as quantum annealing with synthetic data are presented in Figures~\ref{s_fig:lagrange_sim}, \ref{s_fig:lagrange_real}, \ref{s_fig:lagrange_off_sim}, and \ref{s_fig:lagrange_off_real}. In each of the figures, the topmost plot shows the probability of finding the correct solution and the plots below show the measurement energy for increasing noise levels from top to bottom.

\begin{table*}[tb]
\vspace{3pt}
\begin{center}
\small
\input{tables/mot15_all}
\end{center}
\caption{Results on the MOT15~\cite{leal-taixe_motchallenge_2015} training and test set. Results on the training set are generated using leave-one-out cross validation (X-Val).}
\label{tab:mot15_all}
\end{table*}

\paragraph{Fixed Lagrangian.}
Results for a fixed Lagrangian multiplier are shown Figures~\ref{s_fig:lagrange_sim} and \ref{s_fig:lagrange_real} for real and synthetic data respectively. For simulated annealing, the Lagrangian multiplier is in the range $\lambda \in [2, 5]$ and with noise levels between $\sigma=0.2$ to $\sigma=1.0$. For quantum annealing the ranges are $\lambda \in [1, 5]$ and $\sigma \in [0.0, 0.3]$ respectively.

For quantum as well as for simulated annealing, the spectral gap decreases with increasing noise levels, and thus, also the corresponding solution probability. When comparing the two approaches with each other, it becomes apparent that quantum annealing often returns high energy solutions, which corresponds to a higher temperature of the currently available systems.

\paragraph{Optimized Lagrangian.}
Results for optimized Lagrangian multipliers are shown in Figures~\ref{s_fig:lagrange_off_sim} and \ref{s_fig:lagrange_off_real} for real and synthetic data respectively. Noise parameters are the same as for fixed Lagrangian multipliers and the Lagrangian offset $\lambda_\text{off}$ is in the range $\lambda_\text{off} \in [0, 2]$ for both approaches.

Following the same rules as for fixed Lagrangian multipliers, the spectral gap and corresponding solution probability decreases with increasing noise level. Comparing Figure~\ref{s_fig:lagrange_off_sim} to Figure~\ref{s_fig:lagrange_sim} reveals that in simulation a considerable improvement can be achieved by using optimized Lagrangian multipliers. Also for quantum annealing an advantage can be achieved, nevertheless, it is smaller than in simulated annealing, which can be explained by the higher noise level that results in high energy solutions.

\section{MOTChallenge 2015}
\label{s_sec:mot15_details}

Detailed results for our method on each sequence in the MOT15~\cite{leal-taixe_motchallenge_2015} training and test set are provided in Table~\ref{tab:mot15_all}. While the results on both sets are competitive with current state-of-the-art methods~\cite{hornakova_making_2021}, the performance on the training set with leave one out cross-validation is higher than on the test set.

The difference can be explained by the harder examples represented by it. While both splits contain a similar number of frames (5500 frames and 5783 frames respectively), the number of tracks, detected boxes and the corresponding density is approximately $45\%$ higher in the test set and thus, also the complexity and size of the optimization problem. As our formulation is designed for AQC using an Ising model, the resulting optimization problem is a quadratic binary program and thus, hard to solve on classical hardware. This becomes apparent for two sequences in the test set, \textit{AVG-TownCentre} and \textit{PETS09-S2L2} with a high density of $15.9$ and $22.1$ and low frame rate of $2.5\,\text{fps}$ and $7\,\text{fps}$ respectively. The two sequences account for only $27.3\%$ of the total detections, but for $58.7\%$ of the ID switches. ID switches are a good measure for the tracker's performance in this case, as they are less influenced by the performance of the object detector than FP and FN. Due to the larger size of these problems, the optimization cannot finish for all segments within the given time frame and thus, returns a sub-optimal solution.

Even though the problem size is a limitation when solving the problem on classical hardware, it can be resolved when future AQCs become available. As the overall performance is similar to current state-of-the-art methods on MOT15~\cite{leal-taixe_motchallenge_2015}, it can be expected that it scales up to larger datasets accordingly and thus, provides the basis to develop AQC based formulations of the MOT task.

\end{document}

%% file: tables/mot15_short.tex
\setlength{\tabcolsep}{4pt}
\renewcommand\baselinestretch{0.9}\fontsize{7}{9}\selectfont
\begin{tabular}{ll|ccrrrrr}
&Method	&MOTA	&IDF1	&MT	&ML	&FP	&FN	&IDs\\
\midrule
\parbox[t]{1mm}{\multirow{2}{*}{\rotatebox[origin=c]{90}{Test\hspace{18pt}}}}
&Lif\_T~\cite{hornakova_lifted_2020}  &  $\mathbf{52.5}$ & $\mathbf{60.0}$ & $244 $ & $186$ & $6837$ & $21610$ & $730$\\
&MPNTrack~\cite{braso_learning_2020} & $ 51.5 $& $58.6$ & $225$ & $187$ & $7260$ & $21780$ & $\mathbf{375}$ \\ 
&ApLift~\cite{hornakova_making_2021} & $51.1$ & $59.0$ & $\mathbf{284}$ & $\mathbf{163}$ & $10070$ & $\mathbf{19288}$ & $677$\\ 
&MFI\_TST~\cite{yang_online_2021} &$49.2$ &$52.4$ &$210$ &$176$ &$8707$ &$21594$ &$912$\\ 
&Tracktor~\cite{bergmann_tracking_2019} & $ 44.1 $& $46.7$ & $130$ & $189$ & ${6477}$ & $26577$ & $1318$\\
&Ours	&$49.9$	&$53.5$	&$187$	&$179$	&$\mathbf{5924}$	&$24032$	&$1689$\\
\midrule
\parbox[t]{1mm}{\multirow{2}{*}{\rotatebox[origin=c]{90}{X-val}}}
&AP lift	&$59.6$	&$\mathbf{67.8}$	&$\mathbf{237}$	&$\mathbf{133}$	&$8897$	&$\mathbf{10150}$	&$\mathbf{283}$\\
&Ours	&$\mathbf{59.7}$	&$67.6$	&$234$	&$134$	&$\mathbf{8720}$	&$10214$	&$370$\\
\end{tabular}

%% file: tables/mot15_all.tex
\begin{tabular}{ll|ccrrrrr|rrrr}
&seq             &MOTA   &IDF1	&MT	    &ML	    &FP	    &FN	    &IDs   &Density~\cite{leal-taixe_motchallenge_2015}    &Tracks~\cite{leal-taixe_motchallenge_2015}   &Boxes~\cite{leal-taixe_motchallenge_2015}    &FPS~\cite{leal-taixe_motchallenge_2015}\\
\midrule
\parbox[t]{1mm}{\multirow{12}{*}{\rotatebox[origin=c]{90}{X-Val}}}
&Venice-2    	&41.6	&50.0	&13	    &1	    &2178	&1855	&135    &11.9   &26	    &7141   &30\\
&KITTI-17    	&79.6	&83.6	&6  	&0	    &5  	&130	&4      &4.7    &9	    &683    &10\\
&KITTI-13    	&33.5	&57.8	&13 	&11	    &197	&293	&17     &2.2    &42	    &762    &10\\
&ADL-Rundle-8	&26.7	&51.4	&18	    &3  	&3587	&1336	&49     &10.4   &28	    &6783   &30\\
&ADL-Rundle-6	&63.3	&53.7	&11	    &1	    &228	&1570	&40     &9.5    &24	    &5009   &30\\
&ETH-Pedcross2	&46.2	&59.9	&28	    &74	    &127	&3216	&27     &7.5    &133    &6263   &14\\
&ETH-Sunnyday	&78.1	&87.0	&19	    &6	    &110	&295	&2      &5.2    &30	    &1858   &14\\
&ETH-Bahnhof 	&47.3	&67.5	&98	    &38	    &1933	&895	&24     &5.4    &171	&5415   &14\\
&PETS09-S2L1 	&83.2	&76.9	&17	    &0	    &341	&351	&58     &5.6    &19	    &4476   &7\\
&TUD-Campus	    &75.5	&75.4	&4	    &0	    &9	    &72	    &7      &5.1    &8	    &359    &25\\
&TUD-Stadtmitte	&81.6	&80.8	&7  	&0	    &5  	&201	&7      &6.     &10	    &1156   &25\\
&OVERALL	        &59.7	&67.6	&234	&134	&8720	&10214	&370    &7.3   &500   &39905    &-\\
\midrule
\parbox[t]{1mm}{\multirow{12}{*}{\rotatebox[origin=c]{90}{Test}}}
&Venice-1	        &44.4	&49.0	&6	  &3	&656	&1839	&42     &10.1   &17	    &4563   &30\\
&KITTI-19	        &48.2	&60.1	&14	  &17	&528	&2191	&49     &5.0    &62	    &5343   &10\\
&KITTI-16	        &52.7	&67.1	&3	  &1	&120	&666	&19     &8.1    &17	    &1701   &10\\
&ADL-Rundle-3	    &50.0	&47.4	&10	  &7	&653	&4346	&81     &16.3   &44	    &10166  &30\\
&ADL-Rundle-1	    &38.2	&49.9	&12	  &2	&2365	&3313	&73     &18.6   &32	    &9306   &30\\
&AVG-TownCentre	    &52.7	&57.0	&58	  &35	&363	&2767	&250    &15.9   &226	&7148   &2.5\\
&ETH-Crossing	    &62.3	&75.1	&7	  &8	&38	    &335	&5      &4.6    &26	    &1003   &14\\
&ETH-Linthescher    &56.5	&62.3	&45	  &89	&342	&3493	&48     &7.5    &197	&8930   &14\\
&ETH-Jelmoli	    &51.0	&65.5	&18	  &13	&522	&701	&19     &5.8    &45	    &2537   &14\\
&PETS09-S2L2	    &50.1	&38.7	&2	  &4	&312	&4259	&243    &22.1   &42	    &9641   &7\\
&TUD-Crossing	    &85.7	&81.6	&12	  &0	&25	    &122	&11     &5.5    &13	    &1102   &25\\
&OVERALL     	&49.9	&53.5	&187    &179	&5924	&24032	&840 &10.6   &721   &61440       &-\\
\end{tabular}